\documentclass[10pt,twocolumn,letterpaper]{article}

\usepackage[pagenumbers]{cvpr} % To force page numbers, e.g. for an arXiv version

\usepackage[accsupp]{axessibility}  % Improves PDF readability for those with disabilities.

\usepackage{graphicx}
\usepackage{amsmath}
\usepackage{amssymb}
\usepackage{booktabs}
\usepackage{gensymb}

\makeatletter
\@namedef{ver@everyshi.sty}{}
\makeatother
\usepackage{tikz}

\usepackage[pagebackref,breaklinks,colorlinks]{hyperref}

\usepackage[capitalize]{cleveref}
\crefname{section}{Sec.}{Secs.}
\Crefname{section}{Section}{Sections}
\Crefname{table}{Table}{Tables}
\crefname{table}{Tab.}{Tabs.}

\def\cvprPaperID{5644} % *** Enter the CVPR Paper ID here
\def\confName{CVPR}
\def\confYear{2022}

\newcommand{\fakepara}[1]{\vspace{1mm}\noindent\textbf{#1.}}
\DeclareMathAlphabet{\pazocal}{OMS}{zplm}{m}{n}
\newcommand{\unif}{\pazocal{U}}

\newcommand{\bldImageGroup}[1]{\raisebox{-.5\height}{\includegraphics[width=0.19\linewidth]{image/bld/#1/render_pcd.jpg}} &
\raisebox{-.5\height}{\includegraphics[width=0.19\linewidth]{image/bld/#1/render_mesh_poisson.jpg}} &
\raisebox{-.5\height}{\includegraphics[width=0.19\linewidth]{image/bld/#1/render_mesh_neus_nomask.jpg}} &
\raisebox{-.5\height}{\includegraphics[width=0.19\linewidth]{image/bld/#1/render_mesh_siren.jpg}} &
\raisebox{-.5\height}{\includegraphics[width=0.19\linewidth]{image/bld/#1/render_mesh_npsr.jpg}} \\}

\newcommand{\dtuImageGroupSupp}[1]{\vspace{4pt}
	\rotatebox[origin=c]{90}{\small scan #1} &
	\raisebox{-.5\height}{\includegraphics[width=0.19\linewidth]{image/dtu/#1/render_pcd.jpg}} &
	\raisebox{-.5\height}{\includegraphics[width=0.19\linewidth]{image/dtu/#1/render_mesh_poisson.jpg}} &
	\raisebox{-.5\height}{\includegraphics[width=0.19\linewidth]{image/dtu/#1/render_mesh_neus_nomask.jpg}} &
	\raisebox{-.5\height}{\includegraphics[width=0.19\linewidth]{image/dtu/#1/render_mesh_npsr.jpg}} &
	\raisebox{-.5\height}{\includegraphics[width=0.19\linewidth]{image/dtu/#1/render_npsr.jpg}} \\}
\newcommand{\imageGroupSupp}[2]{\vspace{4pt}
	\rotatebox[origin=c]{90}{\small #2} &
	\raisebox{-.5\height}{\includegraphics[width=0.19\linewidth]{image/#1/#2/render_pcd.jpg}} &
	\raisebox{-.5\height}{\includegraphics[width=0.19\linewidth]{image/#1/#2/render_mesh_poisson.jpg}} &
	\raisebox{-.5\height}{\includegraphics[width=0.19\linewidth]{image/#1/#2/render_mesh_neus_nomask.jpg}} &
	\raisebox{-.5\height}{\includegraphics[width=0.19\linewidth]{image/#1/#2/render_mesh_siren.jpg}} &
	\raisebox{-.5\height}{\includegraphics[width=0.19\linewidth]{image/#1/#2/render_mesh_npsr.jpg}} \\}

\begin{document}

\title{Critical Regularizations for Neural Surface Reconstruction in the Wild}

\author{Jingyang Zhang\thanks{This work is done when Jingyang Zhang was an intern at Apple} \qquad Yao Yao \qquad Shiwei Li \qquad Tian Fang \\ David McKinnon \qquad Yanghai Tsin \qquad Long Quan\\
	The Hong Kong University of Science and Technology \qquad Apple
}
\maketitle

%%%%%%%%% ABSTRACT
\begin{abstract}
Neural implicit functions have recently shown promising results on surface reconstructions from multiple views. However, current methods still suffer from excessive time complexity and poor robustness when reconstructing unbounded or complex scenes. In this paper, we present RegSDF, which shows that proper point cloud supervisions and geometry regularizations are sufficient to produce high-quality and robust reconstruction results. Specifically, RegSDF takes an additional oriented point cloud as input, and optimizes a signed distance field and a surface light field within a differentiable rendering framework. We also introduce the two critical regularizations for this optimization. The first one is the Hessian regularization that smoothly diffuses the signed distance values to the entire distance field given noisy and incomplete input. And the second one is the minimal surface regularization that compactly interpolates and extrapolates the missing geometry. Extensive experiments are conducted on DTU, BlendedMVS, and Tanks and Temples datasets. Compared with recent neural surface reconstruction approaches, RegSDF is able to reconstruct surfaces with fine details even for open scenes with complex topologies and unstructured camera trajectories.
\end{abstract}

%%%%%%%%% BODY TEXT
\section{Introduction}\label{sec:intro}

% surface reconstruction by implicit function and differentiable rendering
Surface reconstruction from multiple calibrated views is one of the key tasks in 3D computer vision. Traditionally, the task is solved by first estimating a point cloud from images by multi-view stereo (MVS) \cite{furukawa2009accurate, bleyer2011patchmatch, schonberger2016pixelwise, yao2018mvsnet}, and then extracting a triangular mesh from the point cloud~\cite{labatut2007efficient, kazhdan2013screened, lorensen1987marching}. 
%Recently, neural implicit surface reconstruction also shows comparable results. A direct way is to convert from intermediate representations such as point clouds \cite{sitzmann2020implicit}. Alternatively, the neural surface can be optimized by differentiable rendering \cite{mildenhall2020nerf, niemeyer2020differentiable, yariv2020multiview, zhang2021learning, oechsle2021unisurf, wang2021neus, yariv2021volume}  These methods use multi-layer perceptron (MLP) with space coordinates as input to represent a density, occupancy or signed distance field, as well as scene appearance. Given the geometry and appearance, the systems re-render the input images and optimized by the difference between the rendered images and the input ones. 
Recently, neural implicit surface reconstruction also shows comparable or even better results especially for textureless and non-Lambertian surfaces. These methods apply multi-layer perceptrons (MLP) to map a space coordinate to different geometry properties, such as density \cite{mildenhall2020nerf}, occupancy \cite{niemeyer2020differentiable} or signed distance to the nearest surface point \cite{yariv2020multiview, zhang2021learning, wang2021neus, yariv2021volume}. The MLP can be fit into explicit geometry representations such as contour masks \cite{niemeyer2020differentiable, yariv2020multiview}, depth maps \cite{zhang2021learning} and point clouds \cite{sitzmann2020implicit}, or can be further optimized with the scene appearance through differentiable rendering \cite{niemeyer2020differentiable, yariv2020multiview, zhang2020visibility, mildenhall2020nerf, oechsle2021unisurf, wang2021neus, yariv2021volume}.

% unstructured
However, it is still a challenging task to conduct surface reconstruction in the wild. First, textureless or non-Lambertian surfaces, which exist in real-world scenes, are hard to be recovered even for learning-based methods. Second, camera trajectories of real-world data may be unstructured instead of object-centric.
In traditional mesh reconstruction pipelines, although multi-view stereo \cite{schonberger2016pixelwise, yao2018mvsnet} has been proven to be effective for a variety of different scenes, the reconstructed point cloud inevitably suffers from missing or noisy geometries, which are difficult to be corrected in later mesh extraction steps. On the other hands, recent neural surface methods \cite{mildenhall2020nerf, oechsle2021unisurf, wang2021neus, yariv2021volume} is able to generate surfaces directly from multi-view images. However, the operation of volumetric rendering~\cite{mildenhall2020nerf, oechsle2021unisurf, wang2021neus, yariv2021volume} with implicit functions is time-consuming, and those methods are originally designed for reconstructions of object-centric captures rather than open scenes with unstructured camera trajectories.

In this paper, we propose RegSDF, a neural framework for surface reconstruction from multi-view images for a broad variety of scenes. Implementation-wise, we choose a signed distance field (SDF) as the geometry representation and a surface light field as the appearance model to generate rendered images for network training. To take advantages of well-established MVS pipelines, we additionally apply an oriented point cloud from MVS as input to our reconstruction, where the SDF will be fit into observed data points and also normal directions. Two critical regularizations, namely the Hessian regularization of second derivatives and the minimal surface constraint, are proposed for robust neural surface reconstruction. The Hessian regularization is designed to let the signed distance value smoothly diffuse to the entire signed distance field, which is important for reconstructing complete surface from incomplete and noisy point clouds. Meanwhile, the minimal surface regularization is introduced to compactly interpolate holes and extrapolate missing parts in the implicit surface. We show in experiments that the proposed two regularizations, along with the point cloud supervision, are sufficient to produce high-quality and robust reconstruction results from multi-view images.

The proposed method has been evaluated on \textit{DTU} \cite{jensen2014large}, \textit{BlendedMVS} \cite{yao2020blendedmvs}, and \textit{Tanks and Temples} \cite{knapitsch2017tanks} datasets. We show by both qualitative and quantitative results that our method outperforms other neural implicit surface reconstruction systems by higher surface accuracy, stronger generalization ability to complex scenes, and shorter training time. Also, compared with traditional meshing methods, the proposed framework is robust against point cloud noise and can produce realistic rendered images. 

%-------------------------------------------------------------------------
\section{Related Works}

\fakepara{Differentiable rendering}
Differentiable rendering jointly optimizes all scene parameters including the geometry by inverting the rendering process. In this section we only review neural-based methods. There are two lines of works on the neural differentiable rendering, based on respectively radiance field \cite{mildenhall2020nerf} and surface light field \cite{yariv2020multiview}.

The radiance field representation assumes that radiance is emitted from all space points and applies the volumetric rendering \cite{mildenhall2020nerf, oechsle2021unisurf, wang2021neus, yariv2021volume} to synthesize images. A soft  density field is applied to represent the geometry of the scene. The density value could be further interpreted as occupancy \cite{oechsle2021unisurf} or signed distance \cite{wang2021neus, yariv2021volume} for explicit geometry regularizations. There are two major drawbacks of volumetric rendering based methods. First, it is difficult to accurately extract the surface from the soft density field. Second, the volume rendering requires expensive MLP sampling along the viewing ray.

The surface light field representation follows the assumption that lights are reflected from a opaque geometry surface. For example, DVR \cite{niemeyer2020differentiable} regards the object surface as the zero-crossing interval of the occupancy field, and applies root-finding such as the secant method to find the surface intersection. This process can be accelerated in SDF-based methods \cite{yariv2020multiview, zhang2021learning} by using sphere tracing. However, such methods can only optimize the surface locally, and additional geometric supervisions like masks or depth maps are required to as inputs. As a result, the quality of the surface output will depend on the input data quality. In this paper, we show that proper regularizations, including supervision on second order derivatives and minimal surface constraint, is able to robustly optimize the geometry even for incomplete or missing point cloud inputs.

\fakepara{Neural implicit surface reconstruction}
Recent neural surface reconstruction methods apply MLPs to represent geometries as implicit density field~\cite{mildenhall2020nerf}, occupancy field~\cite{mescheder2019occupancy,saito2019pifu,saito2020pifuhd,peng2020convolutional,niemeyer2020differentiable,oechsle2021unisurf} or signed distance field~\cite{Park_2019_CVPR,liu2019learning,liu2020dist,sitzmann2020implicit,yariv2020multiview, zhang2021learning, wang2021neus, yariv2021volume, kellnhofer2021neural}. The neural surface can be initialized by intermediate geometry representations such as masks, depth maps and point clouds, and can be further optimized with the scene appearance by differentiable rendering. 
%By additional appearance modeling, the systems re-render input views and minimizing the difference between rendered images and input ones. 

Neural implicit surface reconstruction has several advantages over traditional pipelines. First, as the surface is directly modeled and optimized, the final result is optimal with respect to input images, while traditional pipelines produce sub-optimal results through lossy conversions from depth maps to point cloud, and then to triangular meshes. Second, the neural appearance representation can naturally model the view dependent appearance, which is suitable for modeling and reconstructing non-Lambertian surfaces. Inspired by recent works \cite{yariv2020multiview,zhang2021learning}, we apply the SDF and surface light field to represent the geometry and appearance of the scene.

\fakepara{Geometry supervision for neural surface}
Previous works introduce intermediate geometry representations to directly guide the implicit function or provide rough geometric prior. DVR \cite{niemeyer2020differentiable} and IDR \cite{yariv2020multiview} take object masks as inputs to supervise object silhouettes during the network optimization. However, it is difficult to automatically estimate perfect masks for input images, and the silhouette information is not adequate for recovering concave geometries in the scene. MVSDF \cite{zhang2021learning} introduces depth maps from MVS and use SDF fusion to directly supervise the SDF in the whole space, but the method still suffers from incompleteness and noises, making the system unstable for unstructured inputs or complex scenes. In contrast, our methods apply oriented point cloud as inputs, but also takes advantages of input images to refine fine detail geometries in the surface.

\fakepara{Traditional implicit surface reconstruction}
Traditional methods \cite{kazhdan2006poisson, kazhdan2013screened, calakli2011ssd} use volumetric representations like octrees to store  occupancy or signed distance fields in space. Function values and their derivatives are fit to the input data and the Marching Cubes \cite{lorensen1987marching} algorithm is applied to extract surfaces as a level set. These methods can generate triangular meshes with fine details with affordable memory consumption. However, volumetric representation has no intrinsic smoothness compared to neural representation so it is more sensitive to noisy point clouds. Also, octrees are hard to naturally expand to places without observed data points, resulting in missing surfaces when input point clouds are incomplete. Finally, the volumetric representation is hard to model the view-dependent color of the surface, making it hard to optimize for non-Lambertion surfaces. In contrast, neural representation is robust against noisy or missing data, and could be optimized with view-dependent appearance models for realistic view synthesis. 

%-------------------------------------------------------------------------

\section{Method}

\subsection{Geometry and Appearance Representations}\label{sec:geo_and_apr}
Our geometry and appearance modelling, and the construction of differentiable surface intersections closely follow the works in \cite{yariv2020multiview, zhang2021learning}. We define the surface $ \pazocal{S} $ as the zero level set of a Signed Distance Function $ f $ represented by a MLP with network of parameters $ \theta $. Let $ \Omega \subset \mathbb{R}^3 $ be the domain of the scene bounding box. The network function  $ f: \Omega \rightarrow \mathbb{R} $ takes a space location $ \mathbf{x} $ as input and yields the distance from the query location to the nearest surface point. Following the previous level set method \cite{zhao1996variational}, we assume $ f $ is Lipschitz continuous. The scene interior $ \pazocal{V} $  is represented as
\begin{equation*}
	\pazocal{V} = \{ \mathbf{x} \in \Omega \;|\; f(\mathbf{x};\theta)<0 \}, 
\end{equation*}
and the surface $ \pazocal{S} $ as
\begin{equation} \label{eq:levelset}
	\pazocal{S} = \partial \pazocal{V} = \{ \mathbf{x} \in \Omega \;|\; f(\mathbf{x};\theta)=0 \}.
\end{equation}

And the appearance is modeled as a surface light field represented by another MLP $ g $ with network parameters $ \phi $. The network $ g: \pazocal{S} \times \mathbb{R}^3 \times \mathbb{R}^3 \rightarrow [0,1]^3 $ takes an intersection point $ \mathbf{x} $ on surface, the unit normal vector of the point $ \mathbf{n} $, and the unit view direction vector $ \mathbf{v} $ as inputs, and yields the RGB color of this surface point $ \mathbf{c} $: 
\begin{equation*}
	\begin{aligned}
		\mathbf{c}=g(\mathbf{x},\mathbf{n},\mathbf{v};\phi),% \text{, where } \mathbf{x} \in \pazocal{S},
	\end{aligned}
\end{equation*}
where $ \mathbf{x} \in \pazocal{S} $ and the normal can be auto-differentiated from the SDF as $ \mathbf{n} = \nabla_{\mathbf{x}} f(\mathbf{x}; \theta) $. 
%\begin{equation}
%	\mathbf{n} = \nabla_{\mathbf{x}} f(\mathbf{x}; \theta).
%\end{equation}

The intersection points are obtained by sphere tracing algorithm and is not differentiable with respect to the network parameters $ \theta $. To construct a differentiable version $ \mathbf{x}(\theta) $, we can differentiate both sides of $ f(\mathbf{x}(\theta); \theta) = 0 $ with respect to $ \theta $ and rearrange the terms. 
Given the current SDF network parameters $ \theta_0 $, unit view direction vector $ \mathbf{v} $ and the intersection point $ \mathbf{x}_0 $, the differentiable surface intersection $ \mathbf{x}(\theta) $ is derived as:
\begin{equation}
	\begin{aligned}
		\mathbf{x}(\theta) = \mathbf{x}_0 - \frac{f(\mathbf{x}_0;\theta) - f(\mathbf{x}_0;\theta_0)}{\mathbf{v} \cdot \nabla_{\mathbf{x}} f(\mathbf{x}_0;\theta_0)}\mathbf{v},
	\end{aligned}
	\label{eq:intersection}
\end{equation}
where $ f(\mathbf{x}_0;\theta_0) $ and $ \nabla_{\mathbf{x}} f(\mathbf{x}_0;\theta_0) $ are constants. 

\subsection{Data Term}
The proposed method additionally takes as input the oriented point cloud from an MVS network. We first generate point clouds without normal by Vis-MVSNet \cite{zhang2020visibility}, estimate the point cloud normals by principle component analysis for the neighborhood of each point, and align the normal according to camera positions. 

The resulted point cloud can serve as a reliable geometric guidance because they are properly filtered in the MVS step, and interior/exterior of the object can be disambiguated by the normal. 

\fakepara{Surface points}
At the locations of data points, their signed distance are expected to be zero, and surface normals are expected to agree with the data points. Let $ \mathbf{x}_D \in \pazocal{D} $ be all the data points, $ \mathbf{n}_D $ be the unit normal vector of the data points, the data loss is given as 

%\begin{equation}
%	\begin{aligned}
%		L_D = \frac{1}{|\pazocal{D}|} \sum_{\mathbf{x}_D \in \pazocal{D}} \lambda_d |f(\mathbf{x})| + \lambda_n (1 - \frac{\mathbf{n}_D \cdot %\nabla f(\mathbf{\mathbf{x}})}{\| \nabla f(\mathbf{x}) \| } ),
%	\end{aligned}
%\end{equation}

\[
L_D = \frac{1}{|\pazocal{D}|} \sum_{\mathbf{x}_D \in \pazocal{D}} \lambda_d |f(\mathbf{x})| + \lambda_n (1 - \frac{\mathbf{n}_D \cdot \nabla f(\mathbf{\mathbf{x}})}{\| \nabla f(\mathbf{x}) \| } ),
\]
where $ \lambda_d $ and $ \lambda_n $ are the weights for distance term and normal term. 

\begin{figure}
	\centering
	\includegraphics[width=0.8\linewidth]{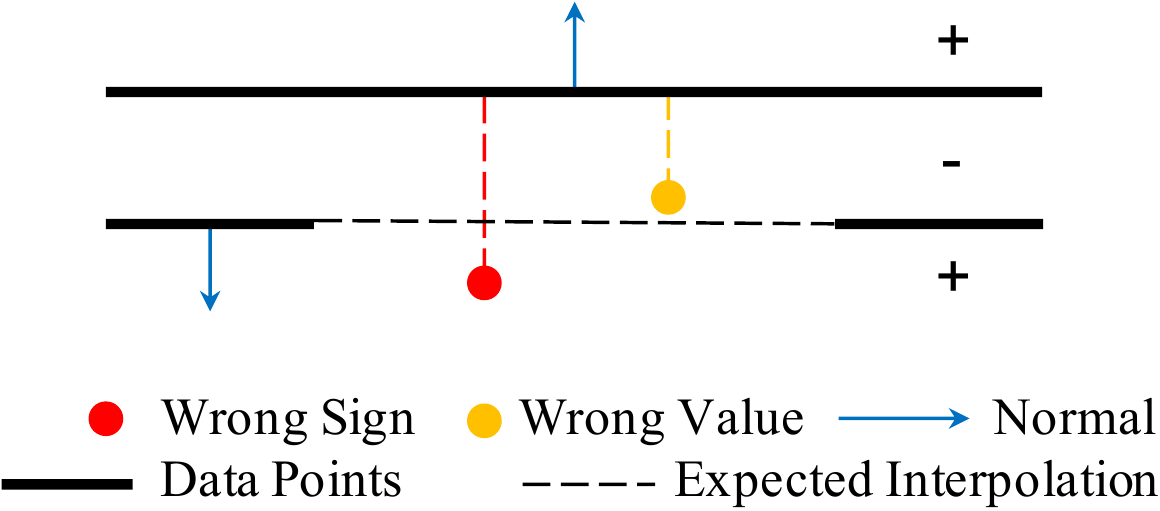}
	\vspace{-3mm}
	\caption{Illustration of wrong estimation of sign and distance value because of the incompleteness in input point clouds, which makes it unreliable to directly supervise the distance in the whole space.}
	\label{fig:wrong_dist}
	\vspace{-3mm}
\end{figure}

\fakepara{Boundary}
We also want to obtain the distance far away from the surface as an additional boundary condition. Camera centers are good choices because cameras must be located in free space and the sign of their distances can be determined. If the camera centers are not inside the bounding box $ \Omega $, we consider the first intersection of the view lines with $ \Omega $. Denote these points as boundary points $ \mathbf{x}_B \in \pazocal{B} $. For each $ \mathbf{x}_B $, we find its nearest neighbor in the point cloud $ \mathbf{x}_B^{nn} \in \pazocal{D} $ with unit normal vector $ \mathbf{n}_B^{nn} $. The distance between $ \mathbf{x}_B $ and its nearest oriented data point $ \mathbf{x}_B^{nn} $ is derived as
\begin{equation*}
	\begin{aligned}
		d(\mathbf{x}_B) = | (\mathbf{x}_B - \mathbf{x}_B^{nn}) \cdot \mathbf{n}_B^{nn} |,
	\end{aligned}
\end{equation*}
and the boundary loss $ L_B $ is the L1 loss between $ f(\mathbf{x}_B) $ and $ d(\mathbf{x}_B) $. 
%In practice, we draw $ k $ nearest neighbors and take the average of each distances. Also, we consider the angle between $ (\mathbf{x}_B - \mathbf{x}_B^{nn}) $ and $ \mathbf{n}_B^{nn} $, and exclude the neighbors with large angle. These two techniques enhance the robustness against missing points and noisy normal. 

Although we can calculate signed distances for any $ \mathbf{x} \in \Omega $ as $ d(\mathbf{x}) = (\mathbf{x} - \mathbf{x}^{nn}) \cdot \mathbf{n}^{nn} $, these distances are unreliable because point clouds from MVS may have missing points so that the real nearest point cannot be found in the input. Fig.~\ref{fig:wrong_dist} shows two examples that distances may have wrong sign or value. However, these problems are not severe for boundary points who have determined sign and are far from the surface. 

\subsection{Regularization}
In this section, we introduce the regularization that smoothly and compactly interpolates and extrapolates distance values away from data points. In each optimization iteration, we uniformly draw samples $ \mathbf{x}_R \in \pazocal{R} \subset \Omega $ from the bounding box and do the following regularization. 

\fakepara{The gradients}
The gradient magnitude is governed by the Eikonal equation $ \| \nabla f(\mathbf{x}) \| = 1 $ in the whole space as in \cite{yariv2020multiview, zhang2021learning}.
The Eikonal loss $L_E$
would therefore expect the gradient magnitude to be 1 in the whole space for the signed distance field without truncation,
it is defined to be summed up over all the random samples and normalized by the number of samples as
\begin{equation*}
	\begin{aligned}
		L_E = \frac{1}{|\pazocal{R}|} \sum_{\mathbf{x}\in \pazocal{R}}| \|\nabla f(\mathbf{x})\| - 1 |.
	\end{aligned}
\end{equation*}

%\fakepara{Direction of Gradients}
The gradient directions are expected not to change rapidly. Following \cite{calakli2011ssd}, we encourage the small second order derivatives everywhere in space and can define a loss using the second order derivatives as Hessian matrix $\mathbf{H}$:
\begin{equation*}
	\begin{aligned}
		L_H = \frac{1}{|\pazocal{R}|} \sum_{\mathbf{x}\in \pazocal{R}} \| \mathbf{H} f(\mathbf{x}) \|_{1},
	\end{aligned}
\end{equation*}
where $ \| \cdot \|_{1} $ is the element-wise matrix 1-norm. 

The smoothness of gradients/normals is also introduced in previous works \cite{oechsle2021unisurf} by minimizing the difference of normals between a surface point and a sample within its neighborhood. Instead of the discrete approach, we calculate analytical Hessian matrices by further back-propagating the gradient computation graph.

This regularization leads to two observations. 
First, the distance values that are properly supervised by the data terms can diffuse to the whole space. As results, we obtain correct distance value away from the surface, and incompleteness of the input can be interpolated. Second, the surfaces with supporting data points are smoothed to avoid overfitting the noise in the input.

% to three results: 1) distance values from the data terms can diffuse to the whole space; 2) surface is smoothed; 3) missing surface in input point %clouds are smoothly interpolated.

\begin{figure}
	\centering
	\includegraphics[width=0.8\linewidth]{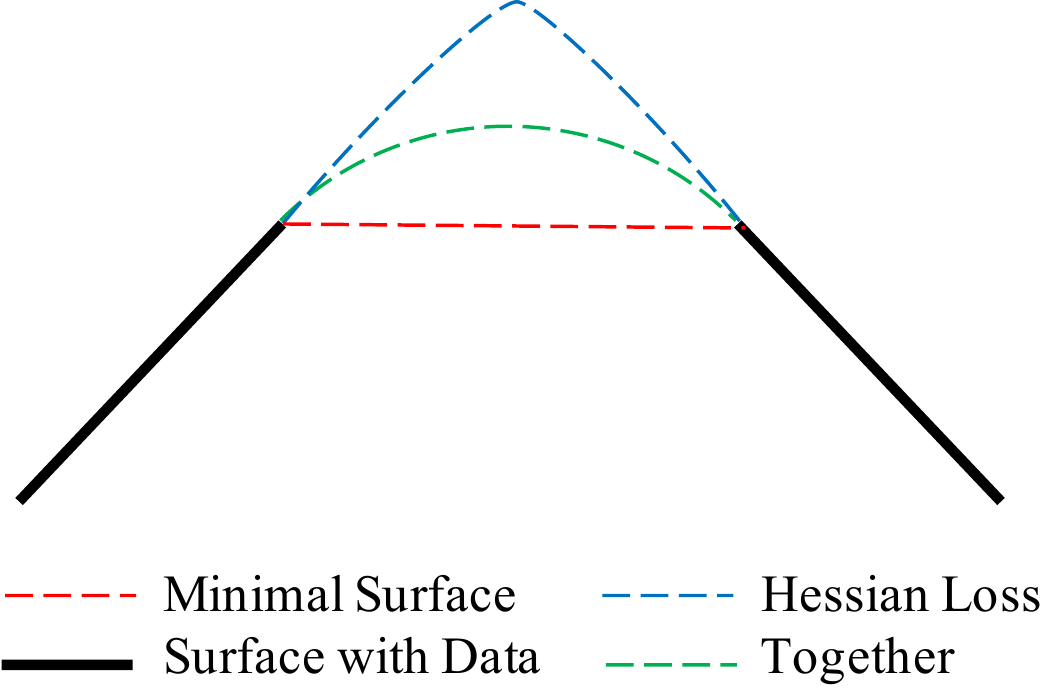}
	\vspace{-3mm}
	\caption{Effects of the regularization. The Hessian loss tends to preserve normals, and the minimal surface constraint closes the surface as planes. We can achieve a natural interpolation by the combination of these two regularizations. }
	\label{fig:regularization}
	\vspace{-3mm}
\end{figure}

\fakepara{The minimal surface}
The surface areas without supporting data points need to be interpolated or extrapolated. The Hessian loss tends to preserve surface normals and extends the existing surfaces as planes.
But when the missing surfaces are non-planar or when we want to 
close a cluster of single-sided point cloud, the extra surfaces may overshoot. 
Moreover, when the scene is not watertight, the surface between object and the boundary of bounding box is random. Inspired by an active contour method in 2D \cite{chan2001active}, we would expect the resulted surface to have minimal total area so that the interpolated and extrapolated surfaces are compact, which is illustrated in Fig.~\ref{fig:regularization}. 

According to Eq.~\ref{eq:levelset}, the volume of the interior can be calculated by
\begin{equation*}
	\text{volume}(\pazocal{V}) = \int_{\Omega} H(f(\mathbf{x}; \theta)) d\mathbf{x}, 
\end{equation*}
area of the implicit surface can be calculated by:
\begin{equation*}
	\begin{aligned}
		%	\text{volume}(\pazocal{V}) &= \int_{\Omega} H(f(\mathbf{x}; \theta)) d\mathbf{x}, \\
		\text{area}(\pazocal{S}) &= \int_{\Omega} \| \nabla H(f(\mathbf{x}; \theta)) \| d\mathbf{x} \\
		&= \int_{\Omega} \delta(f(\mathbf{x}; \theta)) \| \nabla f(\mathbf{x}; \theta) \| d\mathbf{x},
	\end{aligned}
\end{equation*}
where $ H $ is the Heaviside function and $ \delta $ is the Dirac function. Because $ f $ is an SDF, $ \| \nabla f(\mathbf{x}; \theta) \| = 1 $ and can be omitted. In practice, we use a regularized Dirac function $ \delta_\epsilon $ and calculate the integration in a Monte Carlo manner. The minimal surface loss is given by:
\begin{equation*}
	\begin{aligned}
		L_M = \frac{1}{|\pazocal{R}|} \sum_{\mathbf{x}\in \pazocal{R}} \delta_\epsilon (f(\mathbf{x}))\text{, where } \delta_\epsilon(z) = \frac{\epsilon\pi^{-1}}{\epsilon^2+z^2}.
	\end{aligned}
\end{equation*}
We set the parameter $ \epsilon $ controlling the sharpness of the peak to $ \epsilon=10 $ in practice. 

\subsection{Differentiable Rendering}
Similar to recent differentiable rendering methods, we re-render the input images and minimize the difference. In each iteration, we randomly sample pixels $ \mathbf{p} \in \pazocal{I} $ with ground truth color $ \hat{\mathbf{c}}_{\mathbf{p}} $. For each pixel, we first find the surface intersection $ \mathbf{x}^{rt}_{\mathbf{p}} \in \pazocal{S} $ along the view direction $ \mathbf{v}_{\mathbf{p}} $ by sphere tracing. Then we calculate the normal $ \mathbf{n}^{rt}_{\mathbf{p}} = \nabla f(\mathbf{x}) |_{\mathbf{x}=\mathbf{x}^{rt}_{\mathbf{p}}} $ and construct a differentiable intersection $ \mathbf{x}^{rt}_{\mathbf{p}}(\theta) $ by Eq.~\ref{eq:intersection}. Now the color of $ \mathbf{p} $ can be determined by evaluating the surface light field $ g $ with the above inputs and the render loss is given as
\begin{equation*}
	\begin{aligned}
		L_R &= \frac{1}{|\pazocal{I}|} \sum_{\mathbf{p} \in \pazocal{I}} \| {\mathbf{c}}_{\mathbf{p}} - \hat{\mathbf{c}}_{\mathbf{p}} \|_1, \\
		\text{where} \quad {\mathbf{c}}_{\mathbf{p}} &= g( \mathbf{x}^{rt}_{\mathbf{p}}(\theta), \mathbf{n}^{rt}_{\mathbf{p}}, \mathbf{v}_{\mathbf{p}} ).
	\end{aligned}
\end{equation*}
The render loss jointly optimizes the geometry and the appearance. It can recover surface details and refine the interpolated surfaces. 

%-------------------------------------------------------------------------

\begin{table*}[]
	\centering
	\resizebox{\linewidth}{!}{%
		\begin{tabular}{l|cc||cccc|ccc|cc||cc|ccc}
			\specialrule{.2em}{.1em}{.1em}
			& \multicolumn{9}{c|}{Chamfer (mm)}                                    & \multicolumn{7}{c}{PSNR} \\
			\cline{2-17}      & sPSR  & SSD   & NeRF  & UNISURF & NeuS  & VolSDF & IDR   & MVSDF & RegSDF (Ours)  & sPSR  & SSD   & NeRF  & VolSDF & IDR   & MVSDF & RegSDF (Ours) \\
			\hline
			24    & 0.628 & 0.761 & 1.920 & 1.320 & 1.370 & 1.140 & 1.630 & 0.826 & \textbf{0.597} & 19.30 & 19.23 & 26.24 & \textbf{26.28} & 23.29 & 25.02 & 24.78 \\
			37    & 1.335 & 1.657 & 1.730 & 1.360 & \textbf{1.210} & 1.260 & 1.870 & 1.763 & 1.410 & 15.63 & 14.65 & \textbf{25.74} & 25.61 & 21.36 & 19.47 & 23.06 \\
			40    & 0.639 & 0.684 & 1.920 & 1.720 & 0.730 & 0.810 & \textbf{0.630} & 0.883 & 0.637 & 19.51 & 18.49 & \textbf{26.79} & 26.55 & 24.39 & 25.96 & 23.47 \\
			55    & 0.373 & 0.406 & 0.800 & 0.440 & \textbf{0.400} & 0.490 & 0.480 & 0.440 & 0.428 & 19.27 & 19.33 & \textbf{27.57} & 26.76 & 22.96 & 24.14 & 22.21 \\
			63    & 1.061 & 0.888 & 3.410 & 1.350 & 1.200 & 1.250 & \textbf{1.040} & 1.105 & 1.342 & 20.85 & 20.26 & \textbf{31.96} & 31.57 & 23.22 & 22.16 & 28.57 \\
			65    & 0.591 & 0.519 & 1.390 & 0.790 & 0.700 & 0.700 & 0.790 & 0.904 & \textbf{0.623} & 17.72 & 17.83 & \textbf{31.50} & \textbf{31.50} & 23.94 & 26.89 & 25.53 \\
			69    & 0.675 & 0.621 & 1.510 & 0.800 & 0.720 & 0.720 & 0.770 & 0.748 & \textbf{0.599} & 21.65 & 21.81 & \textbf{29.58} & 29.38 & 20.34 & 26.38 & 21.81 \\
			83    & 0.888 & 0.946 & 5.440 & 1.490 & 1.010 & 1.290 & 1.330 & 1.259 & \textbf{0.895} & 23.32 & 23.03 & 32.78 & \textbf{33.23} & 21.87 & 25.79 & 28.89 \\
			97    & 0.862 & 0.694 & 2.040 & 1.370 & 1.160 & 1.180 & 1.160 & 1.018 & \textbf{0.919} & 18.78 & 18.78 & \textbf{28.35} & 28.03 & 22.95 & 26.22 & 26.81 \\
			105   & 0.851 & 0.793 & 1.100 & 0.890 & 0.820 & \textbf{0.700} & 0.760 & 1.347 & 1.020 & 21.65 & 21.60 & 32.08 & \textbf{32.13} & 22.71 & 27.29 & 27.91 \\
			106   & 0.534 & 0.481 & 1.010 & \textbf{0.590} & 0.660 & 0.660 & 0.670 & 0.868 & 0.600 & 21.27 & 21.34 & \textbf{33.49} & 33.16 & 22.81 & 27.78 & 24.71 \\
			110   & 0.811 & 0.744 & 2.880 & 1.470 & 1.690 & 1.080 & 0.900 & 0.844 & \textbf{0.594} & 18.41 & 18.45 & \textbf{31.54} & 31.49 & 21.26 & 23.82 & 25.13 \\
			114   & 0.289 & 0.290 & 0.910 & 0.460 & 0.390 & 0.420 & 0.420 & 0.340 & \textbf{0.297} & 19.56 & 19.58 & \textbf{31.00} & 30.33 & 25.35 & 27.79 & 26.84 \\
			118   & 0.379 & 0.353 & 1.000 & 0.590 & 0.490 & 0.610 & 0.510 & 0.467 & \textbf{0.406} & 23.47 & 23.82 & \textbf{35.59} & 34.90 & 23.54 & 28.60 & 21.67 \\
			122   & 0.413 & 0.503 & 0.790 & 0.620 & 0.510 & 0.550 & 0.530 & 0.465 & \textbf{0.389} & 24.30 & 24.45 & \textbf{35.51} & 34.75 & 27.98 & 31.49 & 28.25 \\
			\hline
			Mean  & 0.688 & 0.689 & 1.857 & 1.017 & 0.871 & 0.857 & 0.899 & 0.885 & \textbf{0.717} & 20.31 & 20.18 & \textbf{30.65} & 30.38 & 23.20 & 25.92 & 25.31 \\
			\specialrule{.2em}{.1em}{.1em}
		\end{tabular}
	}%
	\vspace{-3mm}
	\caption{Quantitative results on DTU \cite{jensen2014large} dataset. Our method achieves the lowest Chamfer distance among the neural surface reconstruction methods and a balance between geometry accuracy and rendering fidelity. *Bold values are the best among the neural methods only.}
	\label{tab:dtu}%
	\vspace{-3mm}
\end{table*}%

\section{Experiments}
\subsection{Implementation} \label{sec:impl}
\fakepara{Point cloud generation}
We use the pre-trained Vis-MVSNet \cite{zhang2020visibility} to generate the point cloud. Hyper-parameters, including number of source views $ N_v $, probability threshold $ \mathbf{p}_t $ and number of views for geometric consistency $ N_f $, are set according to different datasets. We use $ N_v=5$, $N_f=3$, $\mathbf{p}_t=(0.6, 0.7, 0.8) $ for \textit{DTU}, $ N_v=7$, $N_f=3$, $\mathbf{p}_t=(0.6, 0.7, 0.8) $  for \textit{BlendedMVS}, and $ N_v=20$, $N_f=3$, $\mathbf{p}_t=(0.3, 0.4, 0) $ for \textit{Tanks and Temples}. 
Normals of point clouds are calculated by PCA for local point clusters by Open3D \cite{Zhou2018}. Finally point clouds are downsampled to have roughly uniform density, where the target density is selected as the 90\% percentile of the inter-point distance of the original point cloud. 

\begin{figure}
	\centering
	\includegraphics[width=\linewidth,trim={2 2 2 2},clip]{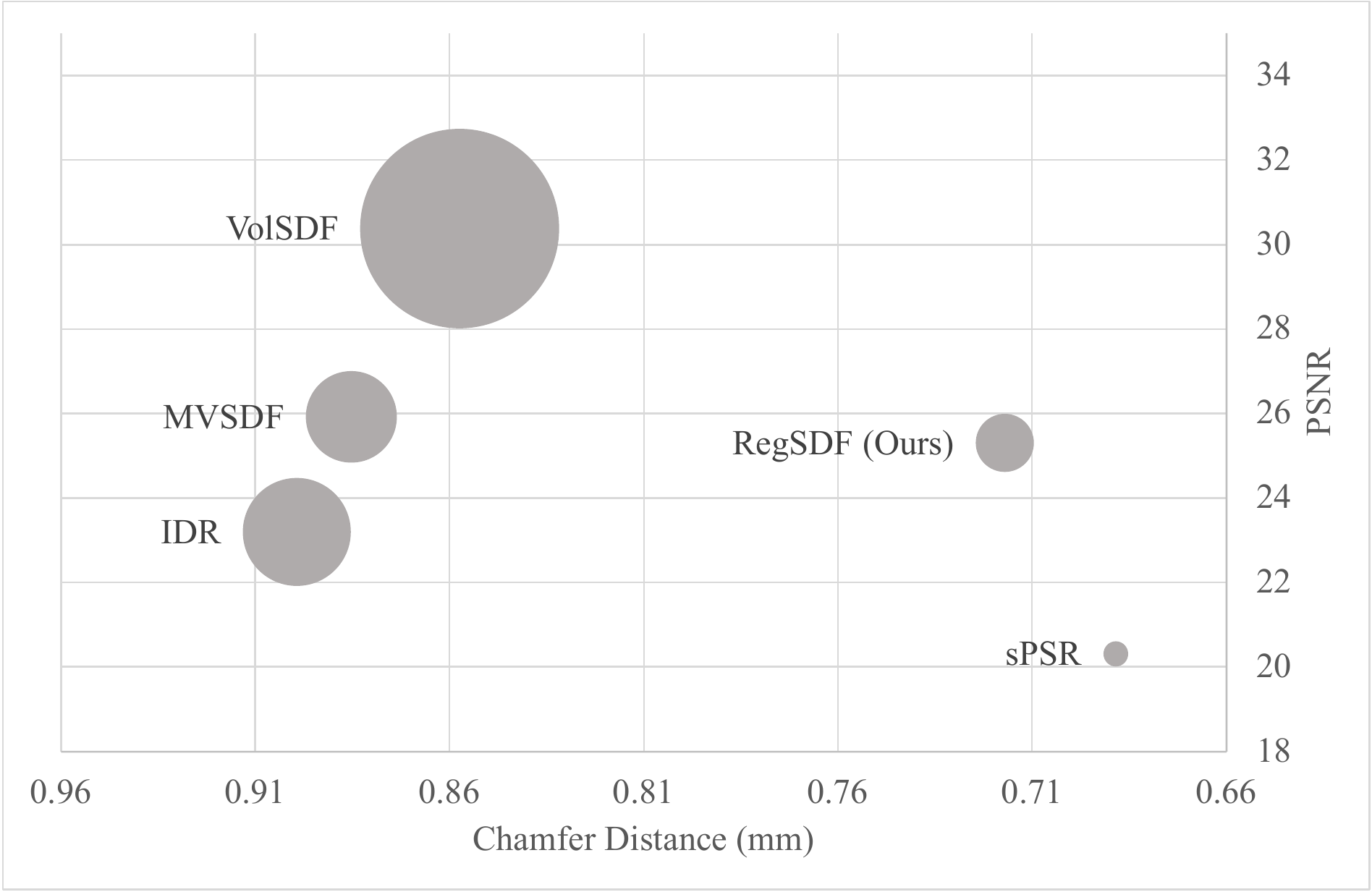}
	\vspace{-8mm}
	\caption{Trade off between geometry accuracy and rendering fidelity. The systems in the upper right corner performs better. The radius of the circles represent the time consumption of each methods. }
	\label{fig:dtu}
	\vspace{-3mm}
\end{figure}

\fakepara{Network architecture}
Following \cite{yariv2020multiview, zhang2021learning}, we use an 8-layer MLP with 512 hidden units and a skip connection in the middle to represent the SDF and a similar 4-layer MLP as the surface light field. Point locations and view directions are enhanced by Positional encoding \cite{mildenhall2020nerf} with 6 and 4 octaves respectively before fed into the networks. The SDF network additionally outputs a 256-channel location descriptor which is then fed into the surface light field. 

\fakepara{Initialization}
Previous methods usually initialize the SDF to roughly form a sphere. However, this initialization is not appropriate for geometries with arbitrary topologies. Instead, we follow \cite{he2015delving} and apply a more general initialization to preserve a uniform distribution $ \unif(-1,1) $ from inputs. 

\fakepara{Loss weights}
The final loss is a weighted sum of the above mentioned losses: $L = \lambda_D L_D + \lambda_B L_B + \lambda_E L_E + \lambda_H L_H + \lambda_M L_M + \lambda_R L_R $. Also, there are also additional weights $\lambda_d $ and $ \lambda_n$ inside $L_D$. If not otherwise specified, we empirically set $ \lambda_d = \lambda_n = \lambda_D = \lambda_B = \lambda_R = 1$ and $L_E = 0.1$, $L_H = L_M = 0.01 $ in our experiments. 

\fakepara{Training}
The network is trained with a batch size of 8 for 1800 epochs for each \textit{DTU} and \textit{BlendedMVS} scene, or 600 epochs for each \textit{Tanks and Temples} scene. The initial learning rate is set to $ 10^{-3} $. Starting from 1/3 of the whole training process, the learning rate is scaled down by $ \sqrt{10} $ every 1/6 of the total epochs. The sample numbers are $ |\pazocal{D}|=32768 $, $ |\pazocal{R}|=16384 $ and $ |\pazocal{I}|=4096\times 8 $. We use the first 1/6 of training as warm up stage where we disable the differentiable intersection to avoid instabilities. 

\begin{figure*}
	\centering
	\begin{tabular}{@{}c@{\hskip2pt}@{\hskip2pt}c@{\hskip2pt}@{\hskip2pt}c@{\hskip2pt}@{\hskip2pt}c@{\hskip2pt}@{\hskip2pt}c@{\hskip2pt}@{\hskip2pt}c@{\hskip2pt}}
		%\dtuImageGroup{24}
		%\dtuImageGroup{37}
		%\dtuImageGroup{40}
		%\dtuImageGroup{65}
		
		%\begin{tikzpicture}\node[above right, inner sep=0](image) at (0,0) {\includegraphics[width=0.3\linewidth]{image/exp2/marna_synp_helmet_ball_combine_city/base_5.jpg}}; \draw[thick,red] (1.1*0.3/0.32,0.3*0.3/0.32) rectangle (2.0*0.3/0.32,1.4*0.3/0.32); \end{tikzpicture}
		
		\vspace{4pt}
		\rotatebox[origin=c]{90}{\small scan 24} &
		\raisebox{-.5\height}{\includegraphics[width=0.19\linewidth]{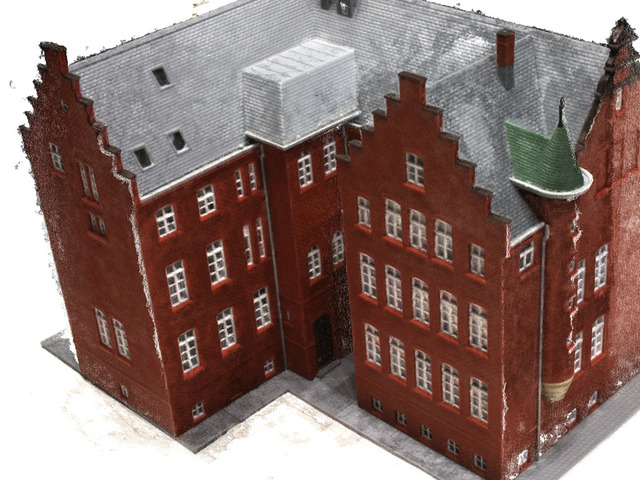}} &
		\raisebox{-.5\height}{\begin{tikzpicture}\node[above right, inner sep=0](image) at (0,0) {\includegraphics[width=0.19\linewidth]{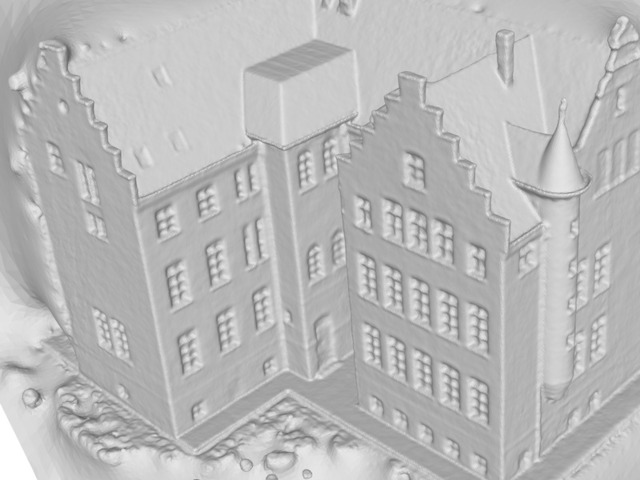}}; \draw[thick,red] (0,0) rectangle (2,0.7); \end{tikzpicture}} &
		\raisebox{-.5\height}{\begin{tikzpicture}\node[above right, inner sep=0](image) at (0,0) {\includegraphics[width=0.19\linewidth]{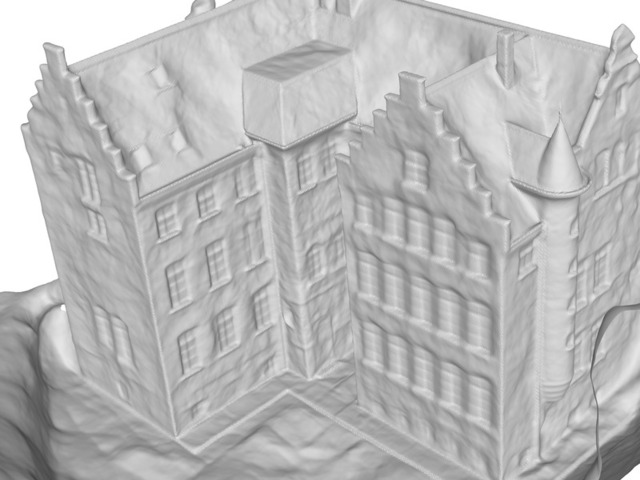}}; \draw[thick,red] (0.7,1.4) rectangle (1.7,2.4); \end{tikzpicture}} &
		\raisebox{-.5\height}{\begin{tikzpicture}\node[above right, inner sep=0](image) at (0,0) {\includegraphics[width=0.19\linewidth]{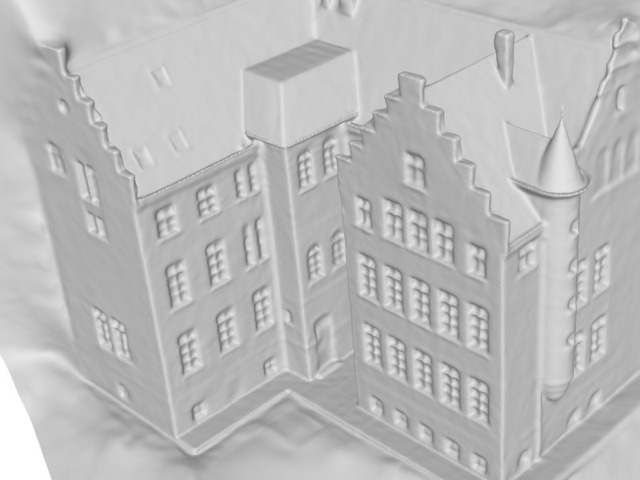}}; \end{tikzpicture}} &
		\raisebox{-.5\height}{\includegraphics[width=0.19\linewidth]{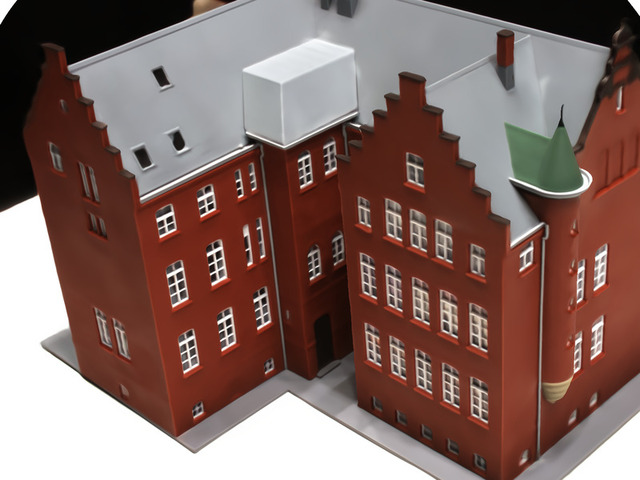}} \\
		
		\vspace{4pt}
		\rotatebox[origin=c]{90}{\small scan 37} &
		\raisebox{-.5\height}{\includegraphics[width=0.19\linewidth]{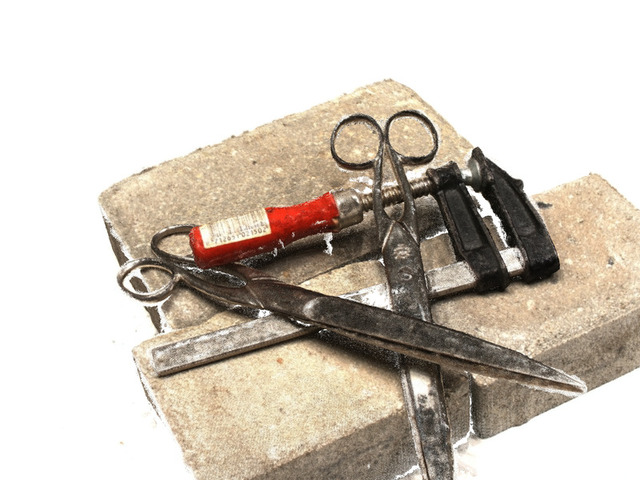}} &
		\raisebox{-.5\height}{\begin{tikzpicture}\node[above right, inner sep=0](image) at (0,0) {\includegraphics[width=0.19\linewidth]{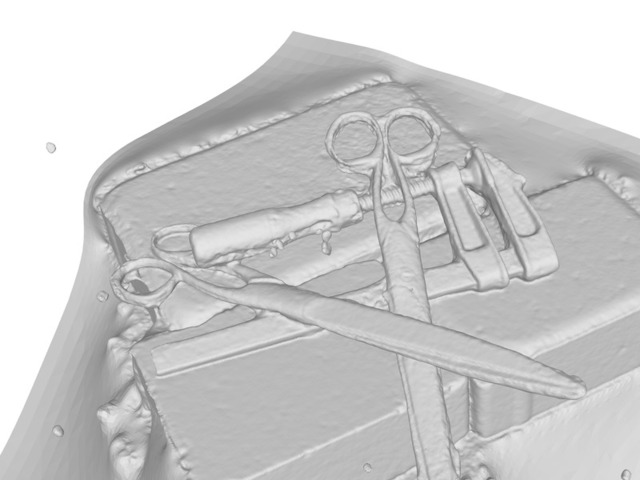}}; \draw[thick,red] (0.9,1.0) rectangle (2,1.6); \end{tikzpicture}} &
		\raisebox{-.5\height}{\begin{tikzpicture}\node[above right, inner sep=0](image) at (0,0) {\includegraphics[width=0.19\linewidth]{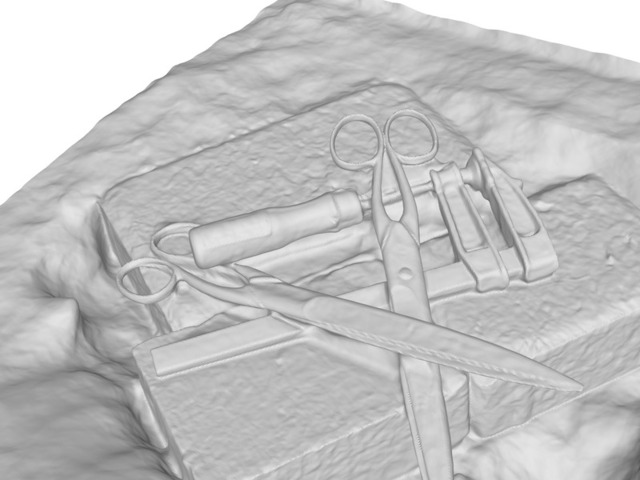}}; \draw[thick,red] (1.2,0.3) rectangle (2.3,1.0); \end{tikzpicture}} &
		\raisebox{-.5\height}{\begin{tikzpicture}\node[above right, inner sep=0](image) at (0,0) {\includegraphics[width=0.19\linewidth]{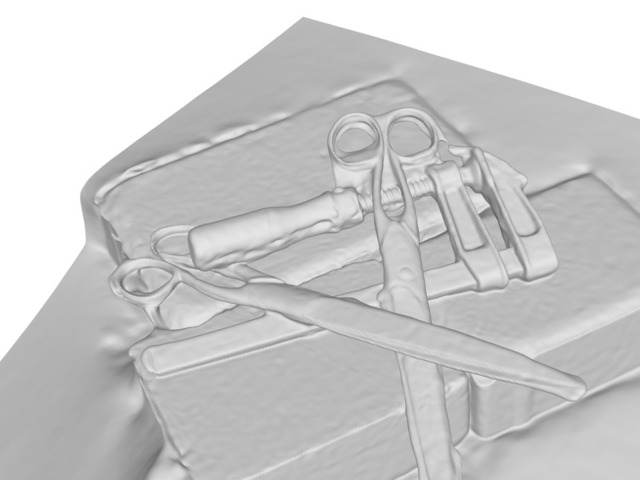}}; \end{tikzpicture}} &
		\raisebox{-.5\height}{\includegraphics[width=0.19\linewidth]{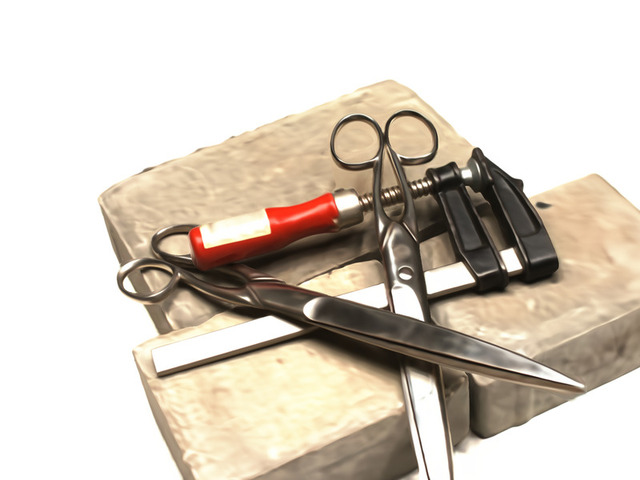}} \\
		
		\vspace{4pt}
		\rotatebox[origin=c]{90}{\small scan 65} &
		\raisebox{-.5\height}{\includegraphics[width=0.19\linewidth]{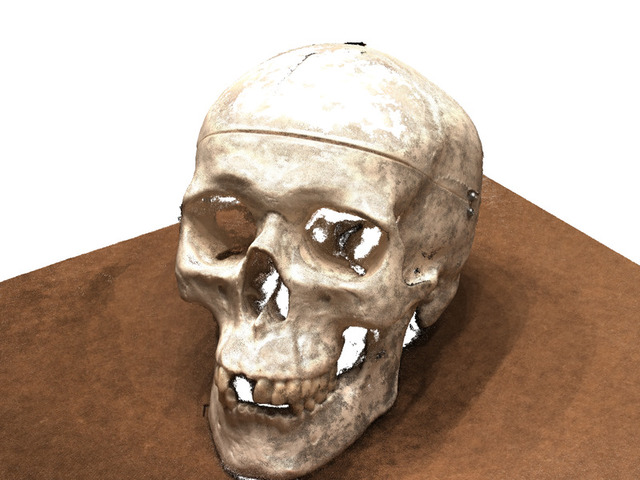}} &
		\raisebox{-.5\height}{\begin{tikzpicture}\node[above right, inner sep=0](image) at (0,0) {\includegraphics[width=0.19\linewidth]{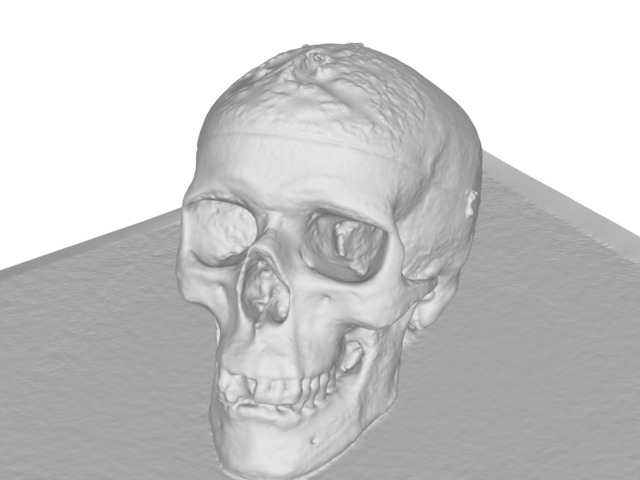}}; \draw[thick,red] (1.0,1.6) rectangle (2.2,2.3); \end{tikzpicture}} &
		\raisebox{-.5\height}{\begin{tikzpicture}\node[above right, inner sep=0](image) at (0,0) {\includegraphics[width=0.19\linewidth]{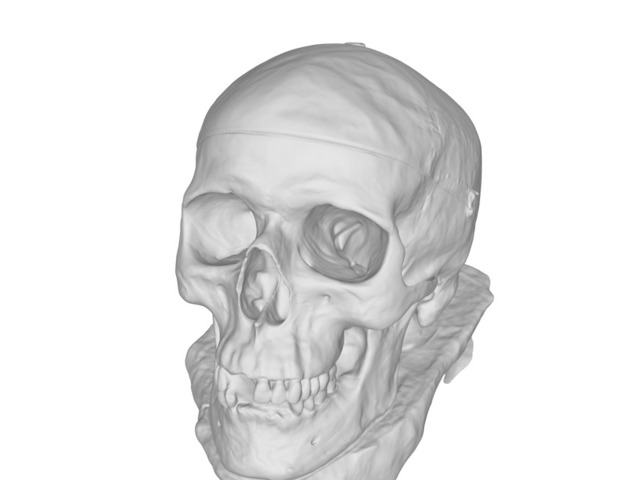}}; \draw[thick,red] (2.0,0) rectangle (3,1); \end{tikzpicture}} &
		\raisebox{-.5\height}{\begin{tikzpicture}\node[above right, inner sep=0](image) at (0,0) {\includegraphics[width=0.19\linewidth]{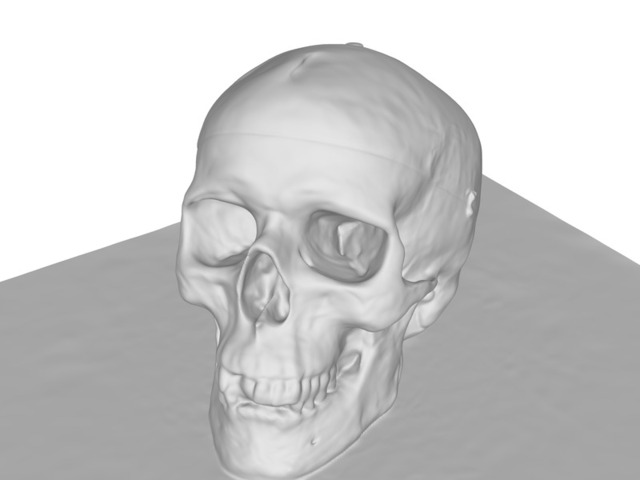}}; \end{tikzpicture}} &
		\raisebox{-.5\height}{\includegraphics[width=0.19\linewidth]{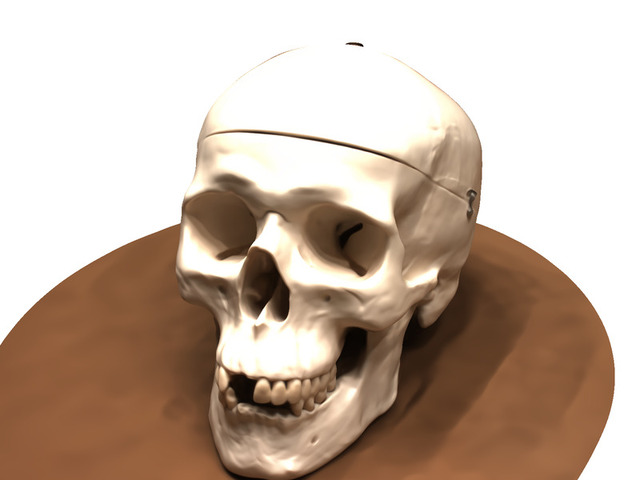}} \\
		
		& Point Cloud & sPSR & NeuS & RegSDF (Ours) & RegSDF Render
	\end{tabular}
	\vspace{-3mm}
	\caption{Qualitative results on DTU \cite{jensen2014large} dataset.}
	\label{fig:dtu_qual}
	\vspace{-3mm}
\end{figure*}
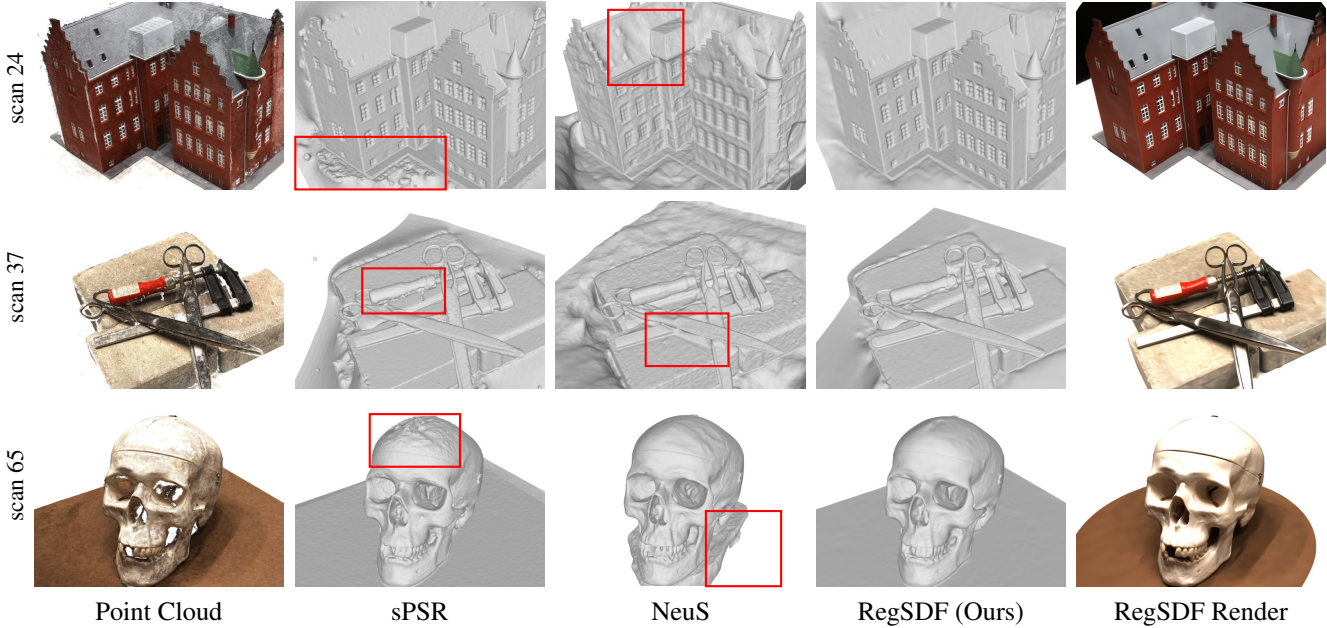

\fakepara{Mesh extraction}
After training, we extract a triangular mesh from the SDF function by Marching Cube \cite{lorensen1987marching} algorithm. The space resolution is set to $1024^3$ for \textit{Tanks and Temples} dataset and open scenes in \textit{BlendedMVS} dataset. For other scenes, the resolution is set to $512^3$. 

\fakepara{Evaluation metrics}
Chamfer distance is a commonly used metric for mesh evaluation. However, we observe that lower Chamfer distance may not necessarily reflects visually better results as the metric only considers the distance between isolated 3D points. To better measure the similarity between meshes, we additionally introduce the surface normal consistency into the metric. Specifically, we first sample oriented point clouds from both meshes where point cloud normals are inherited from mesh triangles. Similar to pure distance metric, we find the nearest neighbor for each point, but additionally consider the normal apart from the distance. A point will be counted as an inlier if and only if the distance and the angular difference between two normals are smaller than a certain threshold. We calculate percentages of inliers in both the estimation (accuracy) and the ground truth (completeness), and report the F-score for evaluating \textit{BlendedMVS} and \textit{Tanks and Temples} datasets. 

\subsection{Baseline Methods}

We compare our method with 1) sPSR \cite{kazhdan2013screened} and SSD \cite{calakli2011ssd}, which are classical mesh reconstructions from point cloud inputs; 2) NeRF \cite{mildenhall2020nerf}, UNISURF \cite{oechsle2021unisurf}, NeuS \cite{wang2021neus} and VolSDF \cite{yariv2021volume}, which are neural surface reconstructions by volumetric neural rendering without additional inputs; 3) IDR \cite{yariv2020multiview} and MVSDF \cite{zhang2021learning}, which are neural surface reconstructions by ray-tracing with additional supervisions. Also, we further compare our method with SIREN \cite{sitzmann2020implicit}, which is also a neural surface reconstruction method from point clouds, on \textit{BlendedMVS} and \textit{Tanks and Temples} datasets. 

\begin{figure*}
	\centering
	\begin{tabular}{@{}c@{\hskip2pt}@{\hskip2pt}c@{\hskip2pt}@{\hskip2pt}c@{\hskip2pt}@{\hskip2pt}c@{\hskip2pt}@{\hskip2pt}c@{\hskip2pt}@{\hskip2pt}c@{\hskip2pt}}
		
		\vspace{4pt}
		\rotatebox[origin=c]{90}{\small building} &
		\bldImageGroup{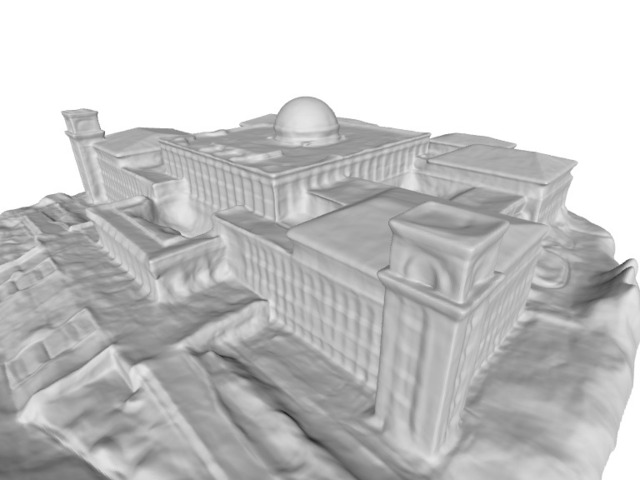}
		
		\vspace{4pt}
		\rotatebox[origin=c]{90}{\small church} &
		\bldImageGroup{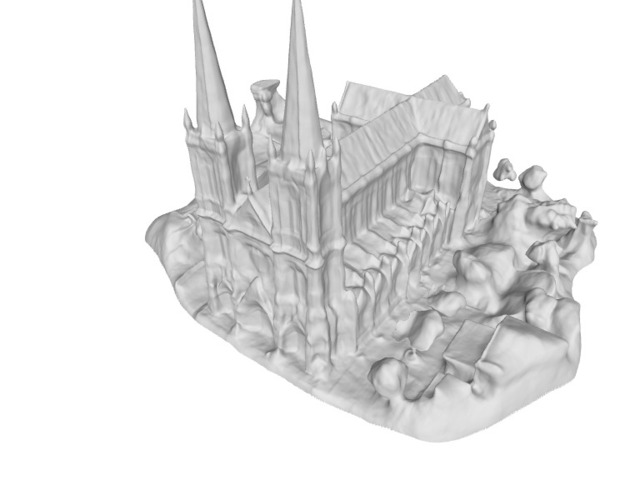}
		
		\vspace{4pt}
		\rotatebox[origin=c]{90}{\small dog} &
		\bldImageGroup{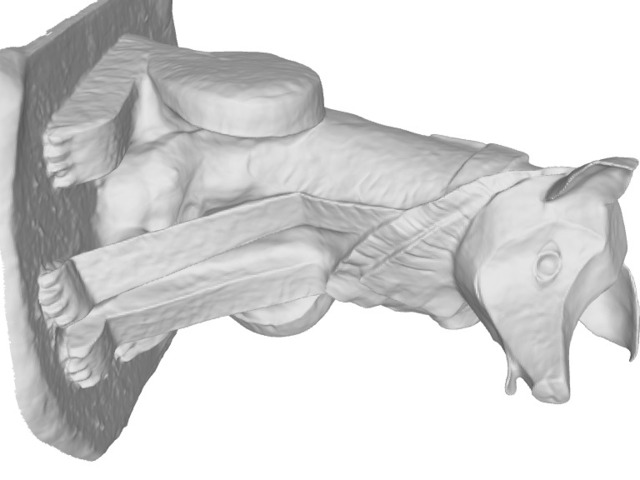}
		
		%\bldImageGroup{jade}
		& Point Cloud & sPSR & NeuS & SIREN & RegSDF (Ours)
	\end{tabular}
	\vspace{-3mm}
	\caption{Qualitative results on BlendedMVS \cite{yao2020blendedmvs} dataset.}
	\label{fig:bld_qual}
	\vspace{-3mm}
\end{figure*}

\begin{figure*}
	\centering
	\begin{tabular}{@{}c@{\hskip2pt}@{\hskip2pt}c@{\hskip2pt}@{\hskip2pt}c@{\hskip2pt}@{\hskip2pt}c@{\hskip2pt}@{\hskip2pt}c@{\hskip2pt}@{\hskip2pt}c@{\hskip2pt}}
		
		\vspace{4pt}
		\rotatebox[origin=c]{90}{\small Barn} &
		\raisebox{-.5\height}{\includegraphics[width=0.19\linewidth]{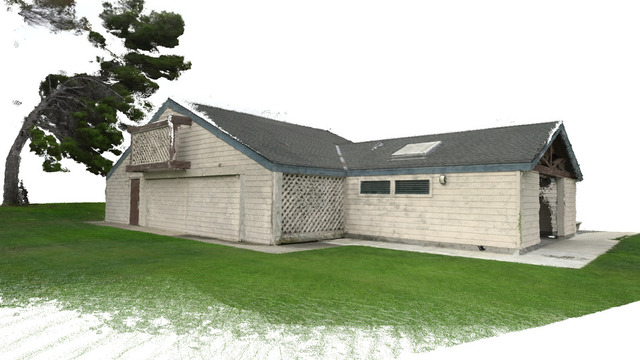}} &
		\raisebox{-.5\height}{\includegraphics[width=0.19\linewidth]{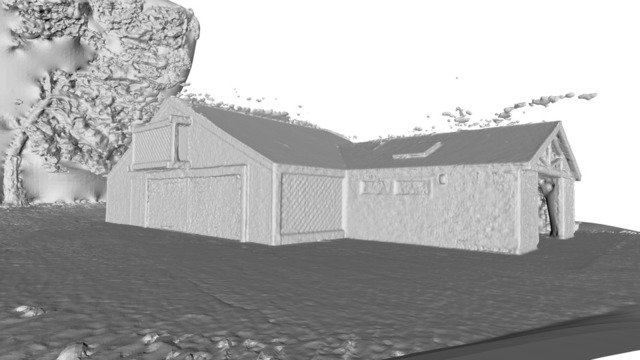}} &
		\raisebox{-.5\height}{\includegraphics[width=0.19\linewidth]{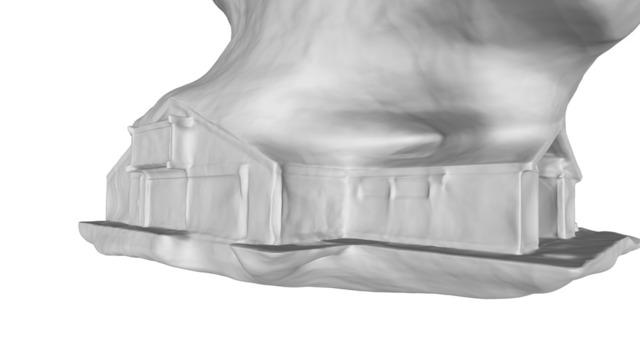}} &
		\raisebox{-.5\height}{\includegraphics[width=0.19\linewidth]{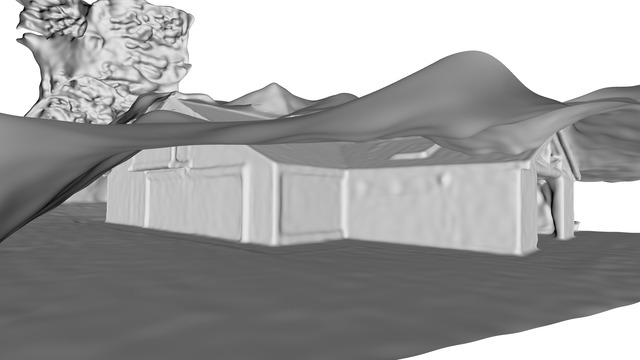}} &
		\raisebox{-.5\height}{\includegraphics[width=0.19\linewidth]{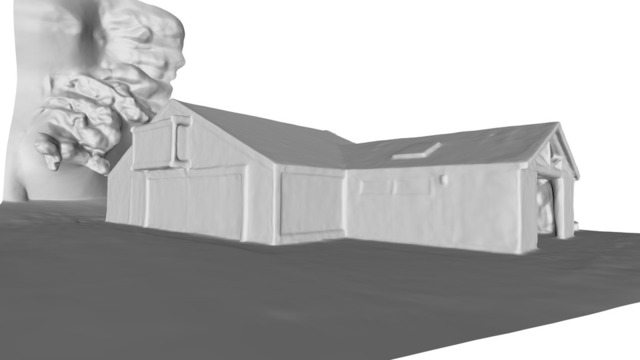}} \\
		
		\vspace{4pt}
		\rotatebox[origin=c]{90}{\small Caterpillar} &
		\raisebox{-.5\height}{\includegraphics[width=0.19\linewidth]{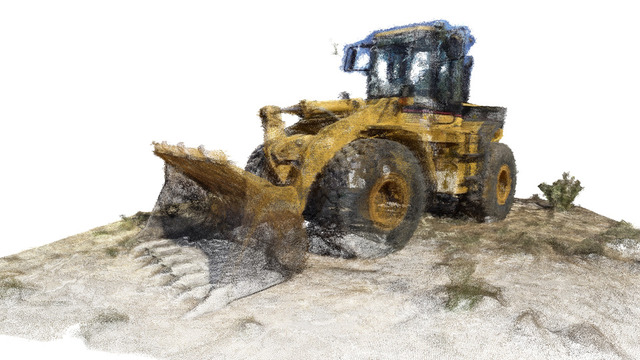}} &
		\raisebox{-.5\height}{\includegraphics[width=0.19\linewidth]{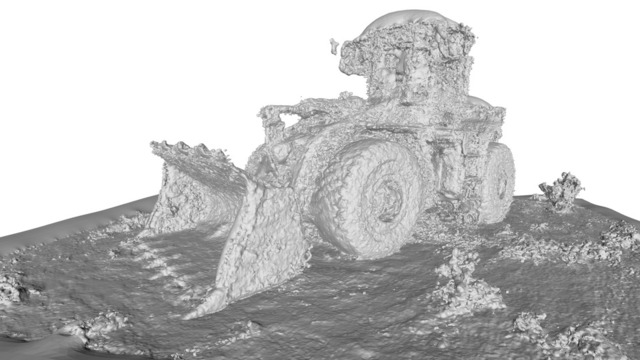}} &
		\raisebox{-.5\height}{\includegraphics[width=0.19\linewidth]{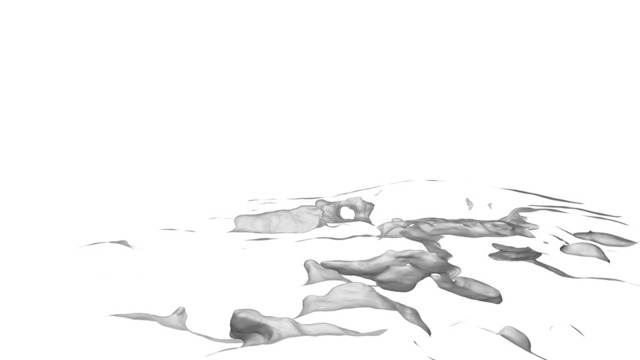}} &
		\raisebox{-.5\height}{\includegraphics[width=0.19\linewidth]{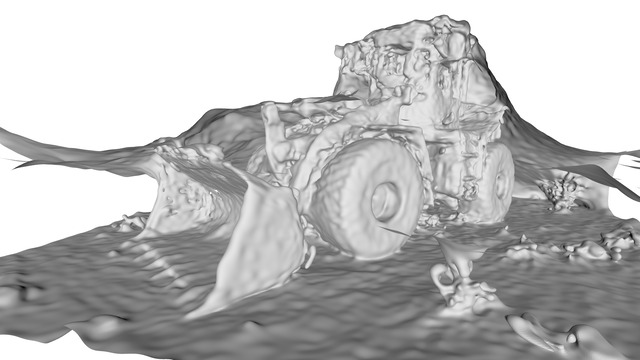}} &
		\raisebox{-.5\height}{\includegraphics[width=0.19\linewidth]{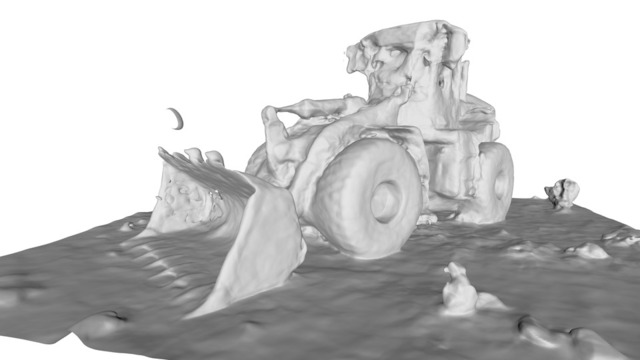}} \\
		
		\vspace{4pt}
		\rotatebox[origin=c]{90}{\small Meetingroom} &
		\raisebox{-.5\height}{\includegraphics[width=0.19\linewidth]{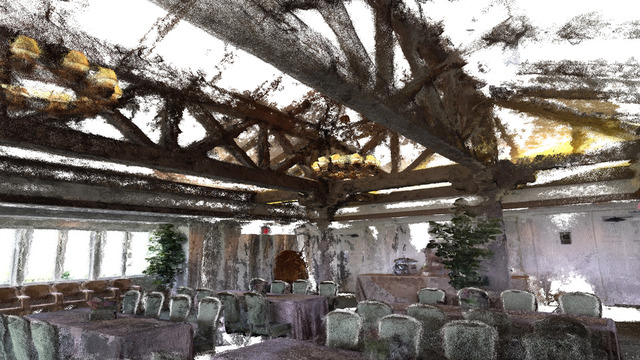}} &
		\raisebox{-.5\height}{\includegraphics[width=0.19\linewidth]{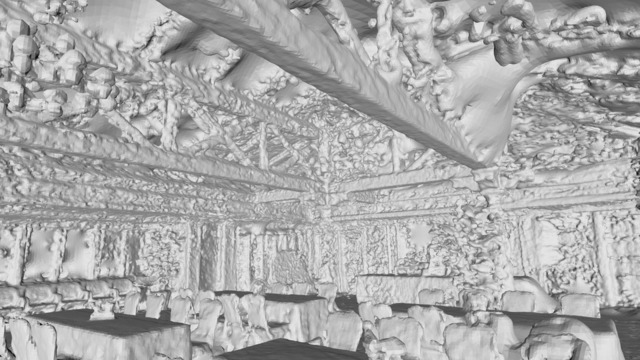}} &
		No Surface &%\includegraphics[width=0.19\linewidth]{image/tnttr/Meetingroom/render_mesh_neus_nomask.jpg} &
		\raisebox{-.5\height}{\includegraphics[width=0.19\linewidth]{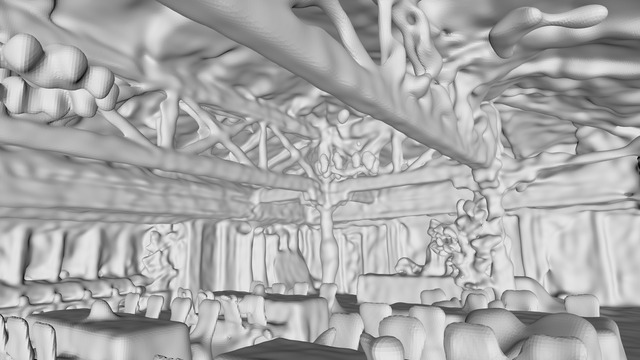}} &
		\raisebox{-.5\height}{\includegraphics[width=0.19\linewidth]{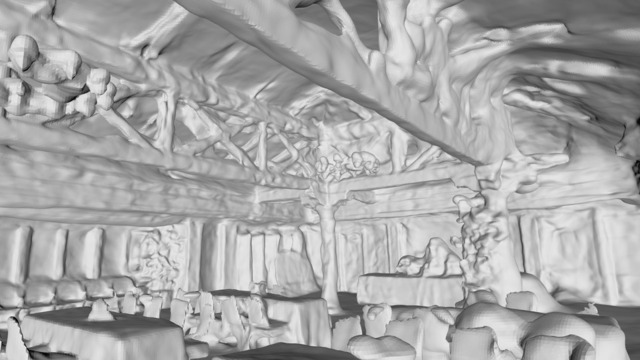}} \\
		
		\vspace{4pt}
		\rotatebox[origin=c]{90}{\small Truck} &
		\raisebox{-.5\height}{\includegraphics[width=0.19\linewidth]{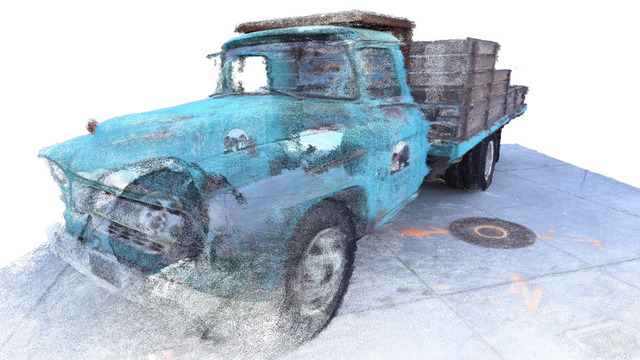}} &
		\raisebox{-.5\height}{\includegraphics[width=0.19\linewidth]{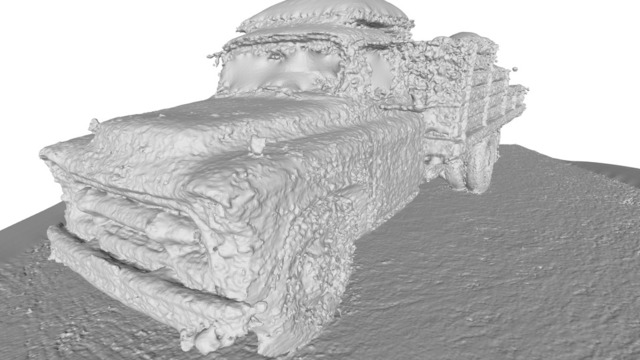}} &
		\raisebox{-.5\height}{\includegraphics[width=0.19\linewidth]{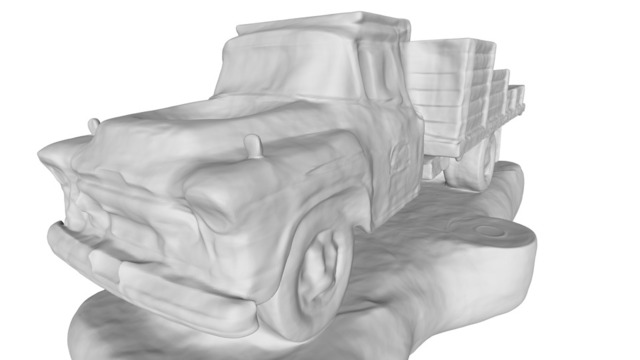}} &
		\raisebox{-.5\height}{\includegraphics[width=0.19\linewidth]{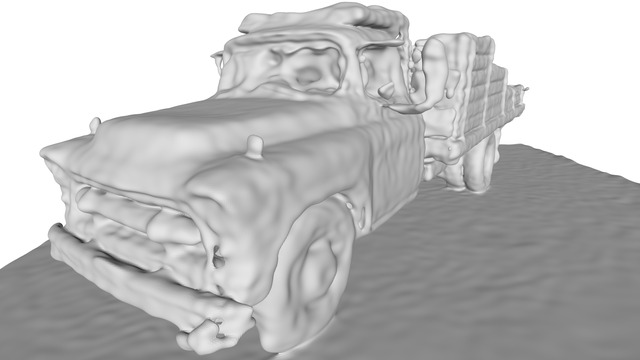}} &
		\raisebox{-.5\height}{\includegraphics[width=0.19\linewidth]{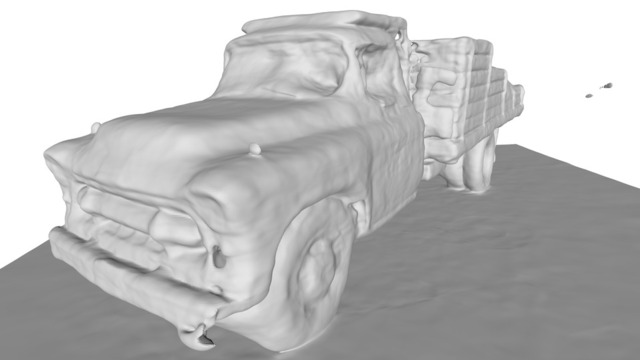}} \\
		
		& Point Cloud & sPSR & NeuS & SIREN & RegSDF (Ours)
	\end{tabular}
	\vspace{-3mm}
	\caption{Qualitative results on Tanks and Temples \cite{knapitsch2017tanks} dataset. SPSR \cite{kazhdan2013screened} produces noisy surface, NeuS \cite{wang2021neus} is not robust for all the scenarios, and SIREN \cite{sitzmann2020implicit} creates incorrect surface closure. }
	\label{fig:tnttr_qual}
	\vspace{-3mm}
\end{figure*}

\subsection{Benchmark on Dataset in Laboratory}

We first benchmark our method on \textit{DTU} dataset~\cite{jensen2014large}. The dataset contains 128 laboratory-captured scenes each with 49 views. All scenes are object-centric and cameras are located in front of the upper bounding sphere of the object. To be consistent with previous methods, we report both surface accuracy as Chamfer distance and render fidelity as PSNR on the same subset of scenes as in \cite{yariv2020multiview, zhang2021learning, oechsle2021unisurf, wang2021neus, yariv2021volume}. 

Comparison on extracted meshes are shown in Fig.~\ref{fig:dtu_qual}. All methods can reconstruct correct topologies of the scenes. It is shown that NeuS produces bumpy surfaces in the rooftop area. We believe it is caused by the ambiguity of disentangling appearance and geometry, which demonstrates the importance of introducing the geometry supervision for neural surface reconstructions. Also, unrestricted surfaces exist in ground areas away from the object, which may be problematic for the reconstruction of non-object-centric scenes.

Quantitative results are shown in Tab.~\ref{tab:dtu}. Our method produce high-quality results similar to classical SSD and sPSR, and is significantly better than recent neural-based methods. Meanwhile, NeRF and VolSDF achieve highest image fidelity scores. In fact, there is a trade off between the geometry accuracy and the rendering fidelity. As shown in Fig.~\ref{fig:dtu}, our method keeps a good balance between the geometry and the appearance qualities. In addition, our method takes less time ($\sim 3.5$ hours) to optimize the scene compared with other neural surface reconstruction methods ($\sim$12 hours for VolSDF).

\subsection{Benchmark on Datasets in the Wild}
\fakepara{BlendedMVS}
Next, we evaluate our method on \textit{BlendedMVS} \cite{yao2020blendedmvs} dataset. We choose to test 4 object-centric scenes and 4 UAV-captured open scenes in the dataset. The distance-only and the normal-aware F-scores described in Sec.~\ref{sec:impl} are used for evaluation. As previous methods have not been evaluated on this dataset, we manually run sPSR, SSD, NeuS and SIREN by provided open-sourced codes. 

Qualitative results are shown in Fig.~\ref{fig:bld_qual}. For NeuS, surfaces near the bounding box boundary are missing. Meanwhile, SIREN tends to produce watertight surfaces, which leads to extra surfaces above or below the scene (see \textit{building} and \textit{church}). Quantitative results are shown in Tab.~\ref{tab:bld}, where our method significantly outperforms all other neural-based methods

\begin{table*}
	\centering
	\resizebox{\linewidth}{!}{%
		\begin{tabular}{l|cc||ccc|cc||ccc}
			\specialrule{.2em}{.1em}{.1em}
			      & \multicolumn{5}{c|}{Distance}         & \multicolumn{5}{c}{Distance and Normal} \\
			\cline{2-11}      & sPSR  & SSD   & NeuS  & SIREN & RegSDF (Ours)  & sPSR  & SSD   & NeuS  & SIREN & RegSDF (Ours) \\
			\hline
			Caterpillar & 23.60\% & 19.99\% & -     & 20.94\% & \textbf{21.54\%} & 13.43\% & 13.16\% & -     & 15.71\% & \textbf{16.75\%} \\
			Truck & 42.64\% & 37.68\% & 24.08\% & 41.93\% & \textbf{46.32\%} & 33.25\% & 31.26\% & 20.19\% & 37.70\% & \textbf{42.03\%} \\
			\hline
			block & 82.04\% & 81.01\% & 48.91\% & 60.43\% & \textbf{77.45\%} & 71.98\% & 71.47\% & 42.03\% & 52.79\% & \textbf{68.06\%} \\
			building & 75.86\% & 75.80\% & 46.89\% & 58.26\% & \textbf{74.50\%} & 61.30\% & 61.78\% & 37.78\% & 46.69\% & \textbf{61.14\%} \\
			church & 70.09\% & 68.01\% & 28.20\% & 56.48\% & \textbf{62.25\%} & 46.66\% & 46.07\% & 16.38\% & 38.97\% & \textbf{43.12\%} \\
			dog   & 72.02\% & 72.68\% & 56.68\% & 68.85\% & \textbf{69.36\%} & 67.88\% & 68.65\% & 52.88\% & 64.53\% & \textbf{65.38\%} \\
			doll  & 83.48\% & 82.36\% & 29.13\% & 59.74\% & \textbf{81.45\%} & 75.71\% & 74.43\% & 22.53\% & 54.20\% & \textbf{74.41\%} \\
			jade  & 53.00\% & 52.83\% & 29.19\% & \textbf{46.47\%} & 43.73\% & 44.58\% & 44.73\% & 22.75\% & \textbf{39.61\%} & 37.21\% \\
			robot & 77.65\% & 68.62\% & 61.33\% & 62.13\% & \textbf{75.04\%} & 70.16\% & 61.99\% & 54.87\% & 56.06\% & \textbf{68.31\%} \\
			ruin  & 73.81\% & 70.06\% & 32.23\% & 58.32\% & \textbf{70.12\%} & 64.59\% & 61.56\% & 27.87\% & 50.81\% & \textbf{62.02\%} \\
			Mean  & 73.49\% & 71.42\% & 41.57\% & 58.84\% & \textbf{69.24\%} & 62.86\% & 61.33\% & 34.64\% & 50.46\% & \textbf{59.96\%} \\
			\specialrule{.2em}{.1em}{.1em}
		\end{tabular}
	}
	\vspace{-3mm}
	\caption{Quantitative results on BlendedMVS \cite{yao2020blendedmvs} and Tanks and Temples \cite{knapitsch2017tanks} dataset. The right side contains the evaluation where both distance and normal similarity are considered as criteria for inliers.}
	\label{tab:bld}%
	\vspace{-3mm}
\end{table*}%

\begin{table*}[h!]
	\centering
	\resizebox{\linewidth}{!}{%
		\begin{tabular}{r|c|ccc|ccc|cc|c}
			\specialrule{.2em}{.1em}{.1em}
			& No    & \multicolumn{3}{c|}{Hessian $ L_H $} & \multicolumn{3}{c|}{Minimal Surface $ L_M $} & \multicolumn{2}{c|}{Sharpness of $ L_M $} & Default \\
			\cline{3-10}      & $L_H$ and $L_M$ & $\lambda_H$=1e-1 & $\lambda_H$=1e-3 & No $L_H$ & $\lambda_M$=1e-1 & $\lambda_M$=1e-3 & No $L_M$ & $\epsilon$=1 & $\epsilon$=100 & Config. \\
			\hline
			DTU & 0.841 & 0.939 & \textbf{0.685} & 0.686 & 0.894 & 0.777 & 0.802 & 0.801 & 0.718 & 0.717 \\
			T\&T.  & 23.90\% & 22.97\% & 28.72\% & 28.51\% & 25.33\% & 29.17\% & 29.25\% & \textbf{29.45\%} & 29.43\% & 29.39\% \\
			\specialrule{.2em}{.1em}{.1em}
		\end{tabular}
	}
	\vspace{-3mm}
	\caption{Ablation and sensitivity study on DTU and Tanks and Temples datasets. The value reported for DTU is the overall Chamfer distance (the lower the better), and for Tanks and Temples is the inlier percentage with normal criterion (the higher the better). The default configuration achieves overall good results. }
	\label{tab:abl_mini}%
	\vspace{-3mm}
\end{table*}%

\begin{figure*}[h!]
	\centering
	\begin{tabular}{@{\hskip2pt}c@{\hskip2pt}@{\hskip2pt}c@{\hskip2pt}@{\hskip2pt}c@{\hskip2pt}@{\hskip2pt}c@{\hskip2pt}@{\hskip2pt}c@{\hskip2pt}}

		\includegraphics[width=0.19\linewidth]{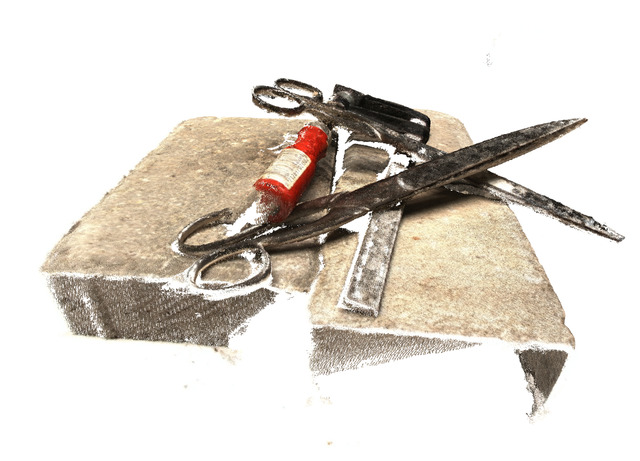} &
		\includegraphics[width=0.19\linewidth]{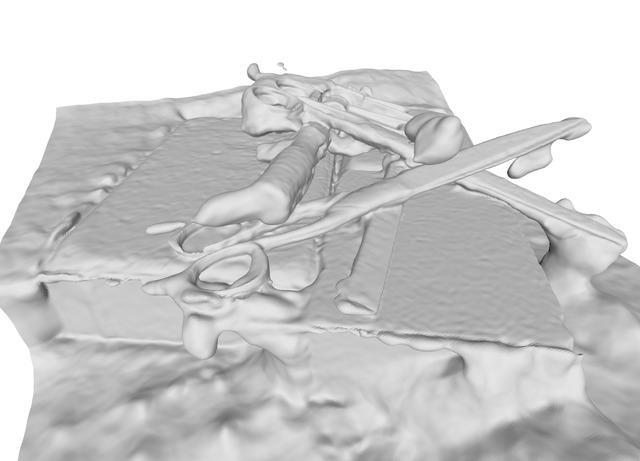} &
		\includegraphics[width=0.19\linewidth]{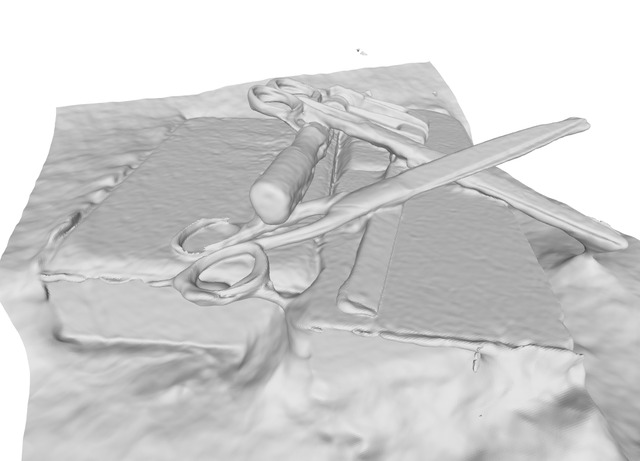} &
		\includegraphics[width=0.19\linewidth]{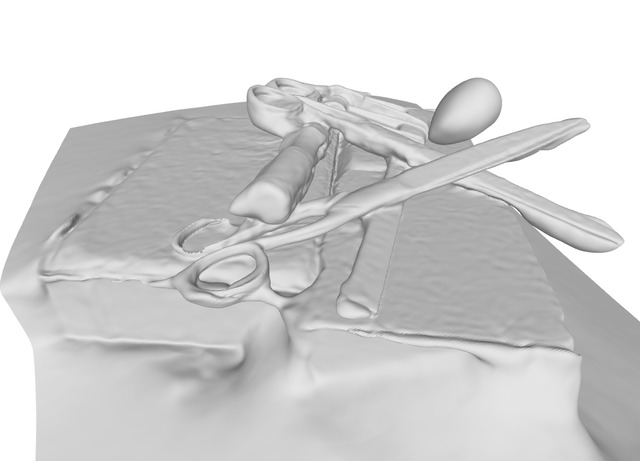} &
		\includegraphics[width=0.19\linewidth]{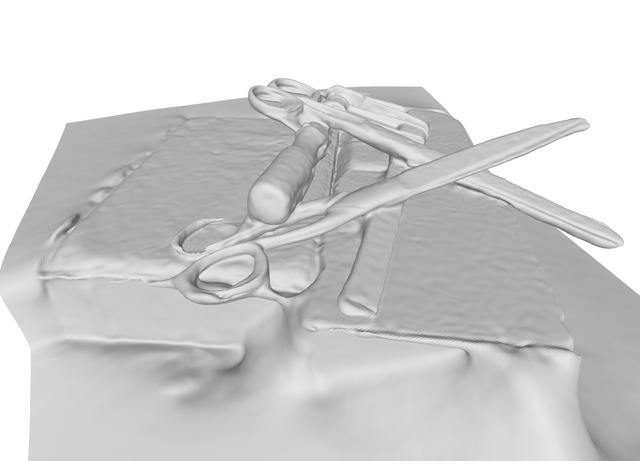} \\
		
		Point Cloud & Remove Both & No Hessian & No Minimal Surface & Default Config. \\
		Chamfer (mm): & 1.780 & 1.516 & 1.600 & 1.410
	\end{tabular}
	\vspace{-3mm}
	\caption{Qualitative results of ablation studies on DTU \cite{jensen2014large} dataset. The Hessian loss produces smooth surface, and the minimal surface constraint creates compact closure of single-sided point cloud (the scissors).}
	\label{fig:dtu_abl}
	\vspace{-3mm}
\end{figure*}

%\subsection{Benchmark on Tanks and Temples}
\fakepara{Tanks and Temples}
Lastly, we test our method on the training set of \textit{Tanks and Temples} dataset \cite{knapitsch2017tanks}. There are totally 7 scenes with 2 indoor scenes which have long been considered challenging for surface reconstruction. We qualitatively compare the 7 scenes and quantitatively evaluate the 2 scenes with ground truth normals (\textit{Caterpillar} and \textit{Truck}). Similar to the evaluation on \textit{BlendedMVS}, we compare our method with sPSR, SSD, NeuS and SIREN using both the distance-only and the normal-aware F-scores. Also, to better adapt NeuS to indoor reconstructions, we modify the SDF geometric initialization by negating the MLP parameters in the last layer. 

Qualitative results are shown in Fig.~\ref{fig:tnttr_qual}. Our method shows the visually best result among all. NeuS fails to reconstruct the indoor scene, and contains inaccurate surfaces for other scenes (\textit{Barn} and \textit{Truck}). SPSR is not robust against noises in input point clouds, which produce noisy and bumpy surfaces in final mesh models. Similar to in BlendedMVS, SIREN produces extra surfaces at scene boundaries, and tends to aggressively close the surface (\textit{Barn} and \textit{Caterpillar}). 
%In contrast, our method can reconstruct smooth surface for various scene topologies.

Quantitative results are shown in Tab.~\ref{tab:bld}. For \textit{Caterpillar}, sPSR produces noisy surface, but quantitatively outperforms our method in terms of Chamfer distance. However, if normal consistency is considered, our method achieves higher score than sPSR among all, which is consistent with qualitative results in Fig.~\ref{fig:tnttr_qual}. 

\subsection{Ablation Study}
We conduct ablation studies to discuss the effectiveness of the Hessian regularization and the minimal surface constraint. We test the following four configurations on DTU and Tanks and Temples datasets: 1) remove both losses; 2) remove Hessian regularization; 3) remove minimal surface constraint; 4) default configuration. Qualitative results of scan 37 in DTU are shown in Fig.~\ref{fig:dtu_abl}. We find that the Hessian regularization can lead to smooth surface, while the minimal surface constraint can avoid extra surface and produce compact closure for single-sided point clouds. Each losses can improve the quantitative evaluation, and the full setting achieves the lowest Chamfer distance among all. 

In addition, we test the system's sensitivity to the loss weights. The quantitative results of ablation and sensitivity study is in Tab.~\ref{tab:abl_mini}. The Hessian loss can be down-weighted  for DTU whose point cloud quality is relatively high, but is essential for Tanks and Temples. Overall, the default configuration achieves good results in all the datasets. 
%For DTU, we can lower the weight of Hessian or even disable it because the point cloud quality is relatively high. 
%For Tanks and Temples, however, changing the weight of Hessian loss do have notable influence on the results. Apart from the Hessian loss, other configurations are worse than or comparable with the default setting. 
%Overall, the default setting achieves good results in all the datasets (which is shown in the main paper). And we believe the default setting will perform well for other new data. A more comprehensive table can be found in the supplementary material. 

%-------------------------------------------------------------------------

\section{Conclusion}
In this work, we proposed RegSDF, which is a novel neural framework for multi-view surface reconstruction. We introduced the MVS point cloud as additional input, and fit the implicit neural surface to the observed 3D points. For rest of the space, we apply the minimal surface constraint and the Hessian constraint on derivatives to regularize the implicit surface. Our method can naturally interpolate the surface where the input point cloud is missing, and is able to mitigate noises in the input point cloud. Extensive evaluations on \textit{DTU}, \textit{BlendedMVS}, and \textit{Tanks and Temples} datasets have shown that the proposed method achieves both accurate surface reconstruction and high-fidelity image rendering for a variety of scenes.

%%%%%%%%% REFERENCES
{\small
	\bibliographystyle{ieee_fullname}
	\bibliography{egbib}
}

%%%%%%%%%%%%%%%%%%%%% supp %%%%%%%%%%%%%%%%%%%%%
\newpage

\twocolumn[
\begin{center}
	\textbf{\LARGE Supplemental Materials}
	\vspace{5mm}
\end{center}
]
\setcounter{section}{0}
\setcounter{equation}{0}
\setcounter{figure}{0}
\setcounter{table}{0}
\setcounter{page}{1}
\makeatletter

\section{Additional Implementation Details}
\fakepara{Distance of Boundary points}
In Sec.~3.2, we calculate nearest neighbor distances $ d(\mathbf{x}_B) = | (\mathbf{x}_B - \mathbf{x}_B^{nn}) \cdot \mathbf{n}_B^{nn} | $ for the boundary points $ \mathbf{x}_B \in \pazocal{B} $, which may suffer from missing geometry or inaccurate estimation of normals. 
In practice, we draw $ k $ nearest neighbors and take the average of each distances. Also, we consider the angle between $ (\mathbf{x}_B - \mathbf{x}_B^{nn}) $ and $ \mathbf{n}_B^{nn} $, and exclude the neighbors with large angle because they are very likely not the nearest neighbors. These two techniques enhance the robustness against missing points and noisy normal. 

\fakepara{Computation of Hessian}
In Sec.~3.3, we introduced that the Hessian matrices can be computed by auto-differentiation, which is similar to computing gradient of the SDF. This process is done by the service provided by Pytorch that a scalar function $ f $ can be differentiated with respect to each input $ (x,y,z) $, and the calculated gradient $ (\frac{\partial f}{\partial x}, \frac{\partial f}{\partial y}, \frac{\partial f}{\partial z}) $ can be added as a part of the computation graph. By further differentiating each entry of the gradient as $ (\nabla \frac{\partial f}{\partial x}, \nabla \frac{\partial f}{\partial y}, \nabla \frac{\partial f}{\partial z}) $ and concatenate the vectors together, we can get the Hessian matrix $ \mathbf{H} f(x,y,z) $. 

\fakepara{Parameter Tuning}
In Sec.~4.1, we introduced the weights of the losses. This configuration can handle most of the cases when the input point cloud does not contain severe noise or incompleteness. Otherwise, we can tune the weights to get better reconstruction. If the geometry is complex and the render loss is not stable, we scale the gradient back-propagated from the differentiable intersection by 0.1. The three scenes \textit{Church}, \textit{Courthouse} and \textit{Meetingroom} of Tanks and Temples dataset \cite{knapitsch2017tanks} are reconstructed under this setting. If the input point cloud is noisy, we change the weight of the normal data term $ \lambda_n $ to 0.1 in the second half of the training process, and let the render loss correct the surface normal. The four scenes \textit{Barn}, \textit{Caterpillar}, \textit{Ignatius} and \textit{Truck} of Tanks and Temples dataset are reconstructed under this setting. 

\section{Baselines}
In this section, we provide more details on how to conduct experiments for the baseline methods on BlendedMVS and Tanks and Temples datasets. 

\fakepara{SPSR and SSD}
We test sPSR \cite{kazhdan2013screened} and SSD \cite{calakli2011ssd} by their official implementation\footnote{https://github.com/mkazhdan/PoissonRecon} and default setting. The resolution is the same as other experiments, which is described in Sec.~4.1. 

\fakepara{NeuS}
We use an unofficial implementation\footnote{https://github.com/ventusff/neurecon} for NeuS \cite{wang2021neus}. The system is trained for a fix number of 300k iterations, which takes 20 hours for each scene. 

\fakepara{SIREN}
We use the official implementation\footnote{https://github.com/vsitzmann/siren} for SIREN \cite{sitzmann2020implicit} The number of training steps is proportional to the number of points in the input point cloud. We add an upper bound of 50k steps. The whole training process takes at most 8 hours for each scene. 

\section{Details of Evaluation Metrics}
In Sec.~4.1, we introduced the metric for evaluating the similarity between meshes, which is used in the comparison on BlendedMVS \cite{yao2020blendedmvs} and Tanks and Temples \cite{knapitsch2017tanks} datasets. Specifically, we first sample points from both meshes with uniform density and preserve their normal. The target density depends on the size of each scene. Then for each point in one point cloud, we find its nearest neighbor in the other one, and check the distance and the normal consistency. The threshold for distances is set to be three times as the downsample density, and the threshold for angular differences is 30\degree. 
%Details of the distance threshold can be found in Tab.~\ref{tab:thresh} %TODO

We run the same evaluation script, which is self-implemented, on both datasets. For Tanks and Temples, our script is different from the official one. First, we only run ICP between the camera trajectories to align the predicted mesh and the ground truth, so the alignments for all the methods are the same. Second, instead of voxel downsampling, we calculate nearest neighbor distances within the point cloud and discard one of the points in the pair with small distance. 

\section{Sensitivity to the Quality of Point Clouds}
\begin{figure*}
	\centering
	\includegraphics[width=\linewidth,trim={2 2 2 2},clip]{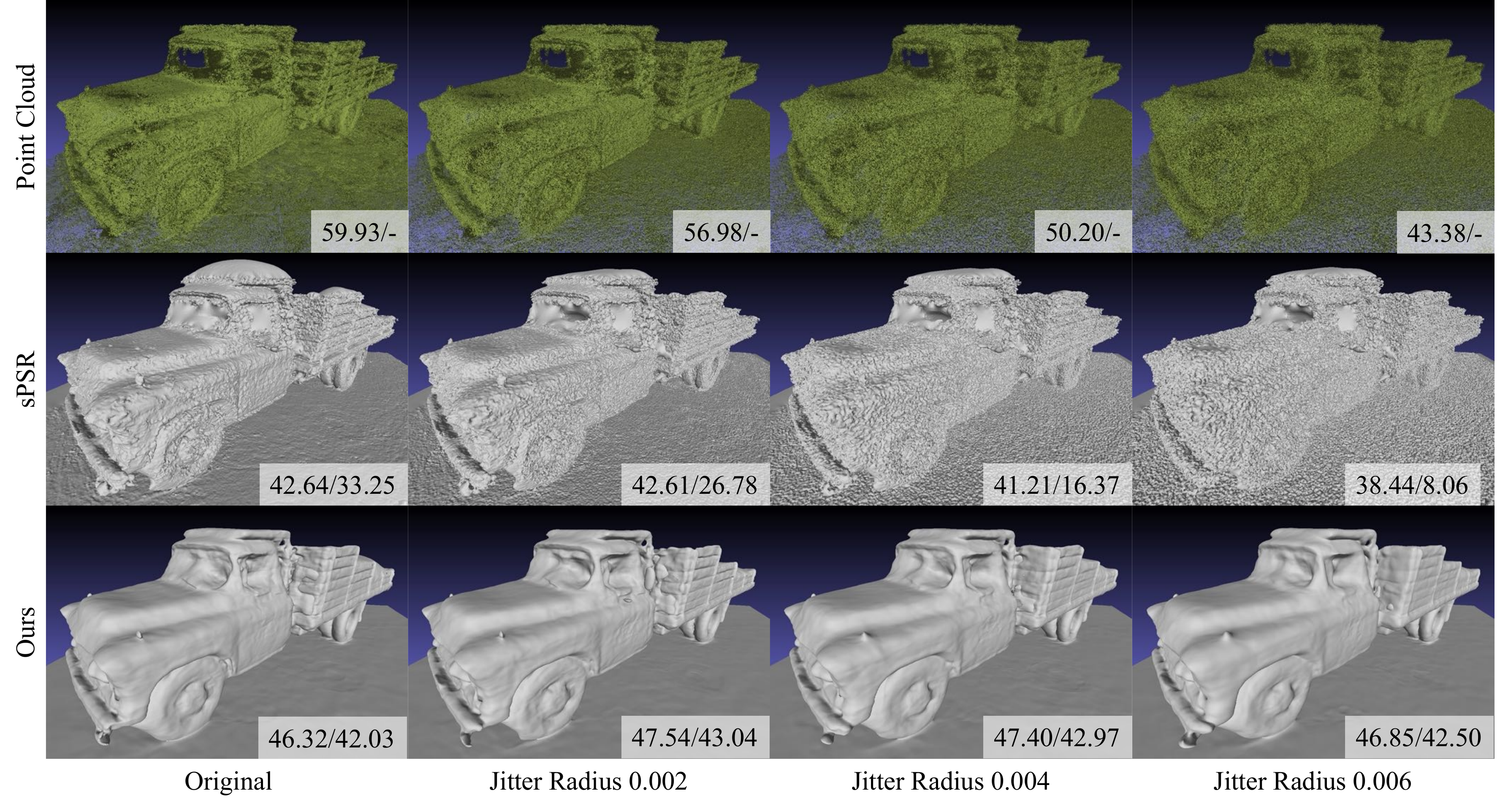}
	\vspace{-3mm}
	\caption{Influence of point cloud quality on mesh accuracy. The two numbers are the evaluation results with and without normal (details are in Sec.~4.1). }
	\label{fig:noise_supp}%
	\vspace{-3mm}
\end{figure*}

We conduct an experiment to examine the sensitivity to the quality of the input point clouds and the results are shown in Fig.~\ref{fig:noise_supp}. Assuming the point cloud from vis-MVSNet is state-of-the-art, we degrade its accuracy by randomly jittering the point positions within a certain radius.

As can be seen, the sPSR method is sensitive as the noise level goes up, but our method is robust against such noises and provides more consistent results both quantitatively and qualitatively. This again supports the argument that proposed regularizations are robust to random noises. %With lower quality of point cloud input it is still able to produce stable quality surface reconstruction.

\section{Ablation and Sensitivity Study of Regularizations}
In Tab.~\ref{tab:abl_supp}, we provide a more comprehensive quantitative ablation and sensitivity study on DTU and Tanks and Temples datasets. 
%In addition, we altered the weights of the two terms and rerun the evaluation to test the sensitivity of the regularizations to the loss weights. The results are also shown in Tab.~\ref{tab:abl_supp}. 
Generally, different hyperparameters do have influence on the result. 
%One empirical observation to note: when the point cloud quality is relatively high (e.g., DTU dataset), we can lower the weight of Hessian or even disable it. 
For DTU, we can lower the weight of Hessian or even disable it because the point cloud quality is relatively high. 
For Tanks and Temples, however, changing the weight of Hessian loss do have notable influence on the results. Apart from the Hessian loss, other configurations are worse than or comparable with the default setting. 
Overall, the default setting achieves good results in all the datasets (which is shown in the main paper). And we believe the default setting will perform well for other new data. 

\begin{table*}
	\centering
	\resizebox{\linewidth}{!}{%
		\begin{tabular}{r|c|ccc|ccc|cc|c}
			\specialrule{.2em}{.1em}{.1em}
			& No    & \multicolumn{3}{c|}{Hessian $ L_H $} & \multicolumn{3}{c|}{Minimal Surface $ L_M $} & \multicolumn{2}{c|}{Sharpness of $ L_M $} & Default \\
			\cline{3-10}      & $L_H$ and $L_M$ & $\lambda_H$=1e-1 & $\lambda_H$=1e-3 & No $L_H$ & $\lambda_M$=1e-1 & $\lambda_M$=1e-3 & No $L_M$ & $\epsilon$=1 & $\epsilon$=100 & Config. \\
			\hline
			24    & 0.808 & 0.777 & 0.586 & \textbf{0.573} & 0.587 & 0.631 & 0.682 & 0.687 & 0.595 & 0.597 \\
			37    & 1.780 & 1.818 & \textbf{1.326} & 1.516 & 1.584 & 1.572 & 1.600 & 1.696 & 1.441 & 1.410 \\
			40    & 0.721 & 0.865 & 0.623 & \textbf{0.610} & 1.132 & 0.659 & 0.675 & 0.671 & 0.633 & 0.637 \\
			55    & 0.480 & 0.483 & \textbf{0.390} & 0.401 & 0.412 & 0.487 & 0.452 & 0.462 & 0.410 & 0.428 \\
			63    & 1.287 & 1.703 & 1.143 & \textbf{1.108} & 1.440 & 1.144 & 1.201 & 1.371 & 1.431 & 1.342 \\
			65    & 0.615 & 1.005 & \textbf{0.611} & 0.620 & 1.575 & 0.742 & 0.742 & 0.673 & 0.632 & 0.623 \\
			69    & 0.765 & 0.771 & 0.576 & \textbf{0.551} & 0.783 & 0.674 & 0.636 & 0.708 & 0.566 & 0.599 \\
			83    & 1.247 & 1.042 & 0.904 & 0.900 & \textbf{0.890} & 1.154 & 1.229 & 1.145 & 0.897 & 0.895 \\
			97    & 0.986 & 1.372 & \textbf{0.848} & 0.866 & 1.206 & 0.909 & 0.895 & 0.948 & 0.942 & 0.919 \\
			105   & 1.324 & 1.196 & 0.976 & \textbf{0.940} & 1.205 & 1.172 & 1.230 & 1.223 & 0.970 & 1.020 \\
			106   & 0.645 & 0.947 & \textbf{0.535} & 0.564 & 0.844 & 0.686 & 0.726 & 0.698 & 0.611 & 0.600 \\
			110   & 0.748 & 0.635 & 0.677 & 0.572 & \textbf{0.537} & 0.720 & 0.786 & 0.673 & 0.580 & 0.594 \\
			114   & 0.335 & 0.360 & \textbf{0.293} & 0.302 & 0.315 & 0.297 & 0.300 & 0.299 & 0.302 & 0.297 \\
			118   & 0.438 & 0.591 & 0.398 & 0.392 & 0.513 & 0.373 & 0.406 & \textbf{0.370} & 0.376 & 0.406 \\
			122   & 0.443 & 0.525 & 0.383 & 0.379 & 0.387 & 0.438 & 0.465 & 0.394 & \textbf{0.379} & 0.389 \\
			\hline
			Mean & 0.841 & 0.939 & \textbf{0.685} & 0.686 & 0.894 & 0.777 & 0.802 & 0.801 & 0.718 & 0.717 \\
			\specialrule{.2em}{.1em}{.1em}
			Caterpillar & 14.33\% & 9.93\% & 16.67\% & 16.47\% & 14.26\% & 16.92\% & 16.64\% & 16.92\% & \textbf{17.62\%} & 16.75\% \\
			Truck & 33.47\% & 36.01\% & 40.78\% & 40.55\% & 36.40\% & 41.42\% & 41.86\% & 41.99\% & 41.24\% & \textbf{42.03\%} \\
			\hline
			Mean  & 23.90\% & 22.97\% & 28.72\% & 28.51\% & 25.33\% & 29.17\% & 29.25\% & \textbf{29.45\%} & 29.43\% & 29.39\% \\
			\specialrule{.2em}{.1em}{.1em}
		\end{tabular}
	}
	\vspace{-3mm}
	\caption{Ablation and sensitivity study on DTU and Tanks and Temples datasets. The value reported for DTU is the overall Chamfer distance (the lower the better), and for Tanks and Temples is the inlier percentage with normal criterion (the higher the better). The default configuration achieves overall good results. }
	\label{tab:abl_supp}%
	\vspace{-3mm}
\end{table*}%

%\section{Training Progression}
\section{More Results}
We provide more results of DTU \cite{jensen2014large}, BlendedMVS \cite{yao2020blendedmvs} and Tanks and Temples \cite{knapitsch2017tanks} datasets in Fig.~\ref{fig:dtu_qual_supp}, \ref{fig:dtu_qual2_supp}, \ref{fig:bld_qual_supp}. 

In the supplementary video, we show novel view rendering results, and the geometry with respect to training steps. 

\begin{figure*}
	\centering
	\begin{tabular}{@{}c@{\hskip2pt}@{\hskip2pt}c@{\hskip2pt}@{\hskip2pt}c@{\hskip2pt}@{\hskip2pt}c@{\hskip2pt}@{\hskip2pt}c@{\hskip2pt}@{\hskip2pt}c@{\hskip2pt}}
		\dtuImageGroupSupp{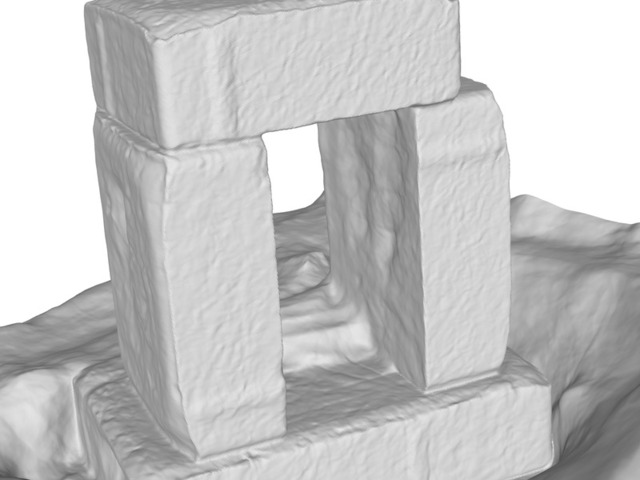}
		\dtuImageGroupSupp{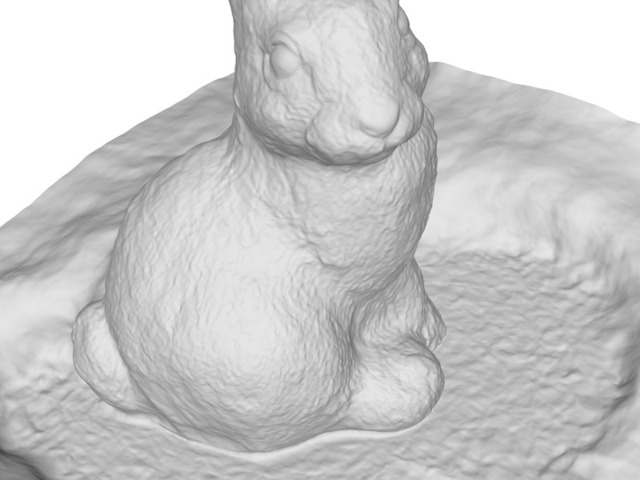}
		\dtuImageGroupSupp{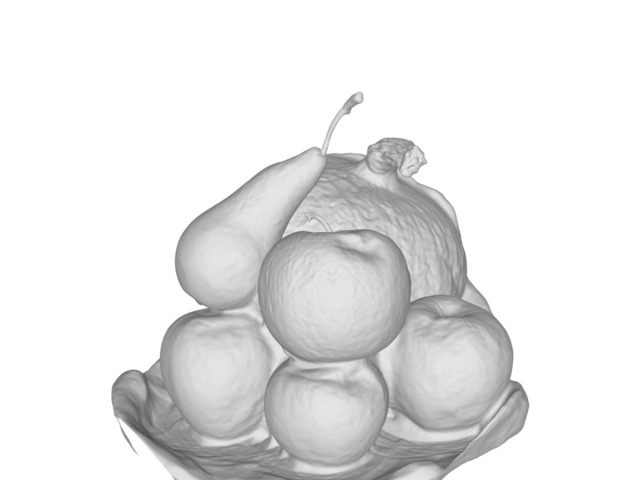}
		\dtuImageGroupSupp{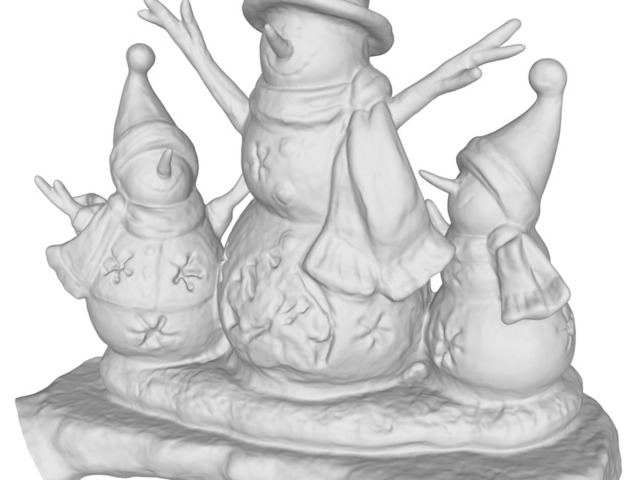}
		\dtuImageGroupSupp{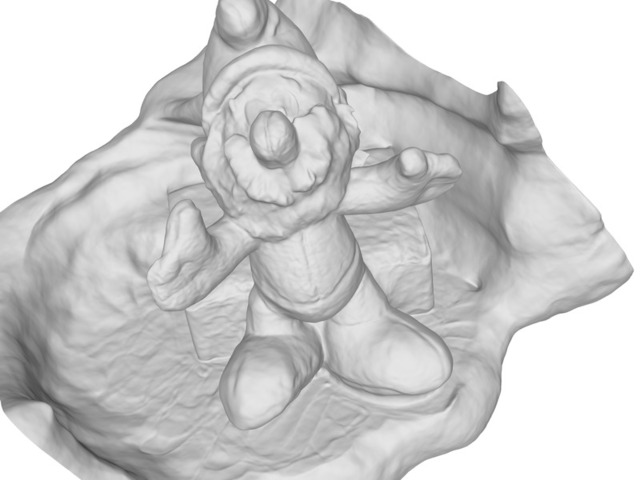}
		\dtuImageGroupSupp{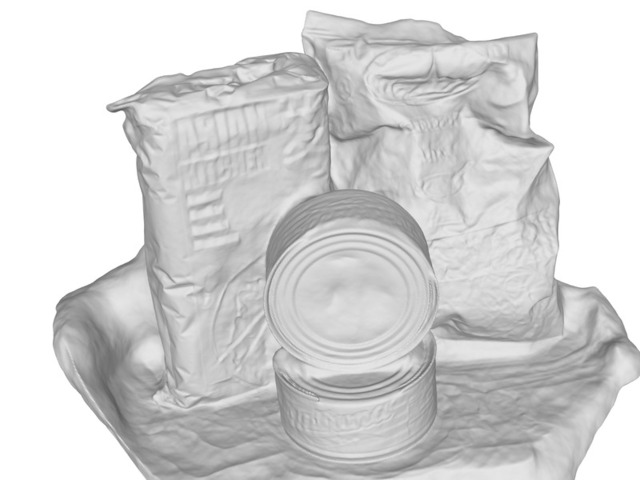}
		& Point Cloud & sPSR & NeuS & RegSDF (Ours) & RegSDF Render
	\end{tabular}
	\vspace{-3mm}
	\caption{Qualitative results on DTU \cite{jensen2014large} dataset. Brand names are blurred. }
	\label{fig:dtu_qual_supp}
	\vspace{-3mm}
\end{figure*}

\begin{figure*}
	\centering
	\begin{tabular}{@{}c@{\hskip2pt}@{\hskip2pt}c@{\hskip2pt}@{\hskip2pt}c@{\hskip2pt}@{\hskip2pt}c@{\hskip2pt}@{\hskip2pt}c@{\hskip2pt}@{\hskip2pt}c@{\hskip2pt}}
		\dtuImageGroupSupp{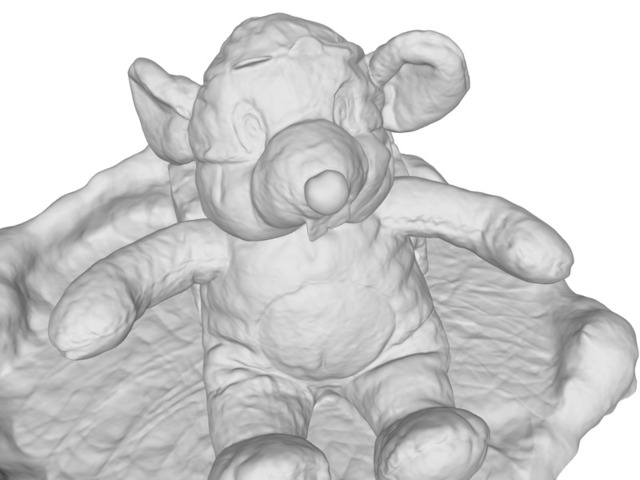}
		\dtuImageGroupSupp{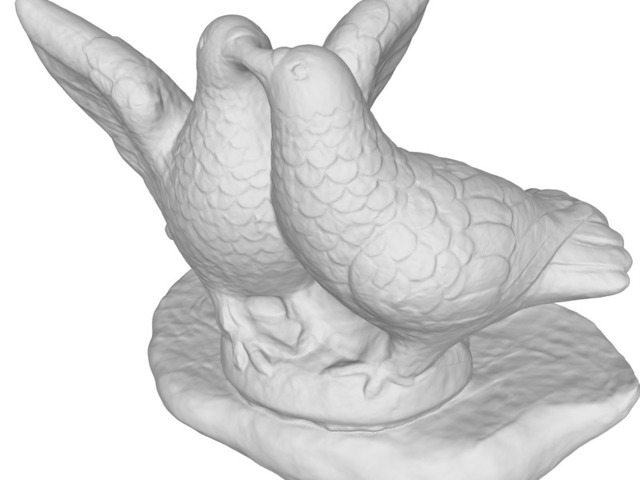}
		\dtuImageGroupSupp{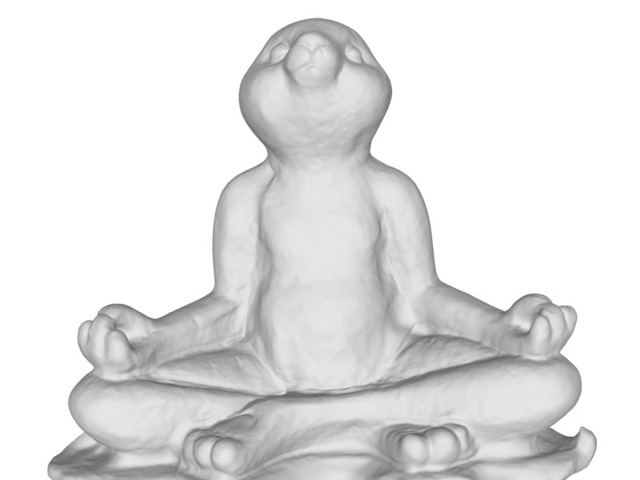}
		\dtuImageGroupSupp{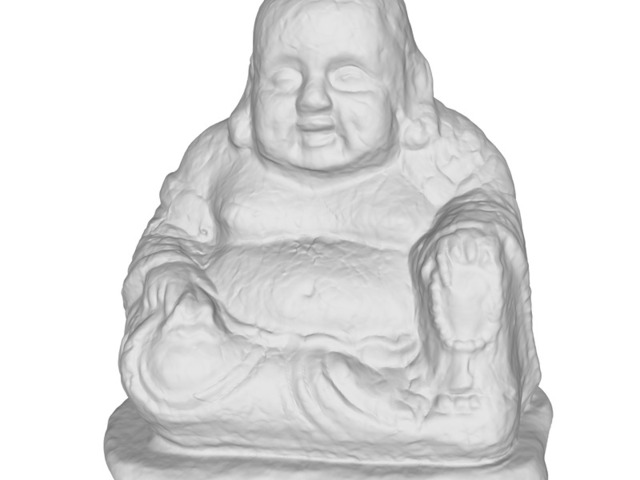}
		\dtuImageGroupSupp{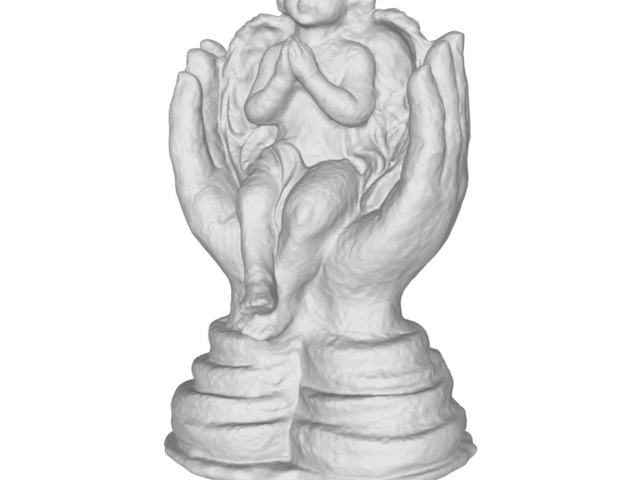}
		\dtuImageGroupSupp{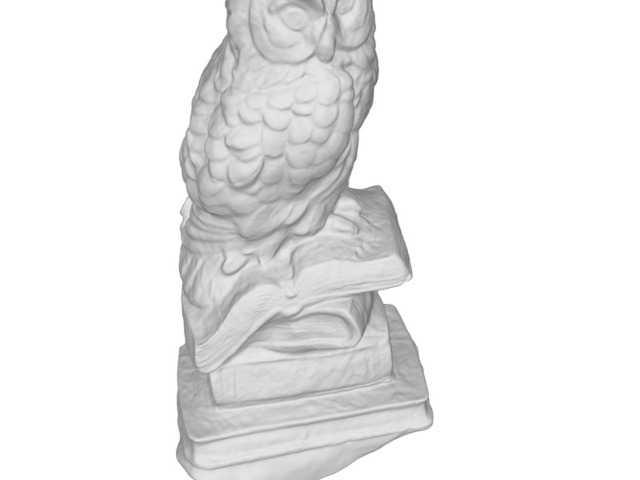}
		& Point Cloud & sPSR & NeuS & RegSDF (Ours) & RegSDF Render
	\end{tabular}
	\vspace{-3mm}
	\caption{Qualitative results on DTU \cite{jensen2014large} dataset (cont.).}
	\label{fig:dtu_qual2_supp}
	\vspace{-3mm}
\end{figure*}

\begin{figure*}
	\centering
	\begin{tabular}{@{}c@{\hskip2pt}@{\hskip2pt}c@{\hskip2pt}@{\hskip2pt}c@{\hskip2pt}@{\hskip2pt}c@{\hskip2pt}@{\hskip2pt}c@{\hskip2pt}@{\hskip2pt}c@{\hskip2pt}}
		\imageGroupSupp{bld}{block}
		\imageGroupSupp{bld}{doll}
		\imageGroupSupp{bld}{jade}
		\imageGroupSupp{bld}{robot}
		\imageGroupSupp{bld}{ruin}
		
		\vspace{4pt}
		\rotatebox[origin=c]{90}{\small Church} &
		\raisebox{-.5\height}{\includegraphics[width=0.19\linewidth]{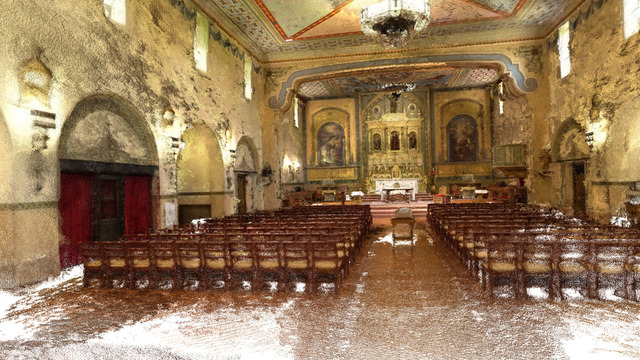}} &
		\raisebox{-.5\height}{\includegraphics[width=0.19\linewidth]{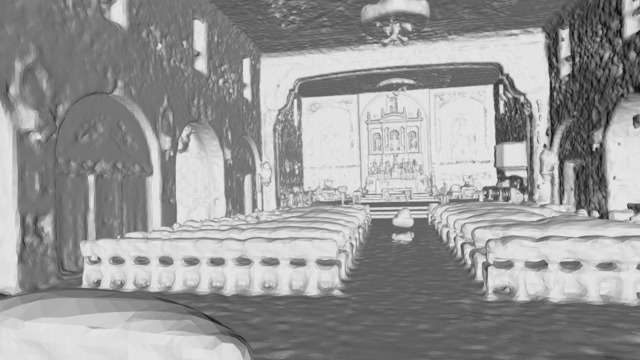}} &
		\raisebox{-.5\height}{\includegraphics[width=0.19\linewidth]{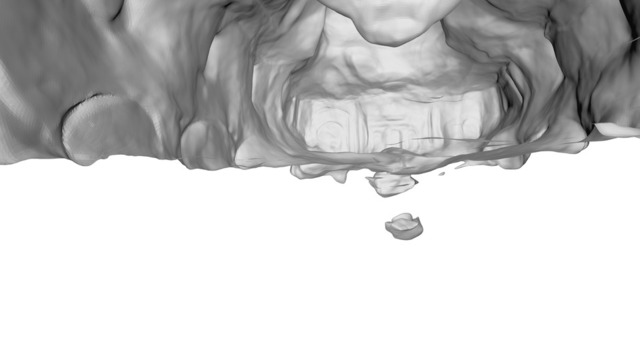}} &
		\raisebox{-.5\height}{\includegraphics[width=0.19\linewidth]{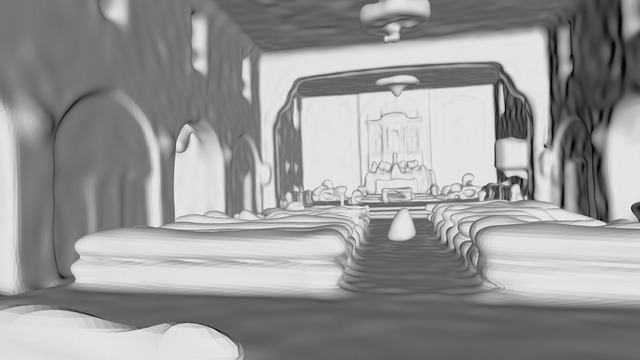}} &
		\raisebox{-.5\height}{\includegraphics[width=0.19\linewidth]{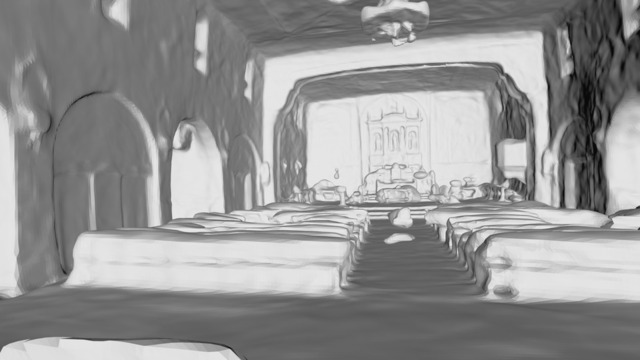}} \\
		
		\vspace{4pt}
		\rotatebox[origin=c]{90}{\small Courthouse} &
		\raisebox{-.5\height}{\includegraphics[width=0.19\linewidth]{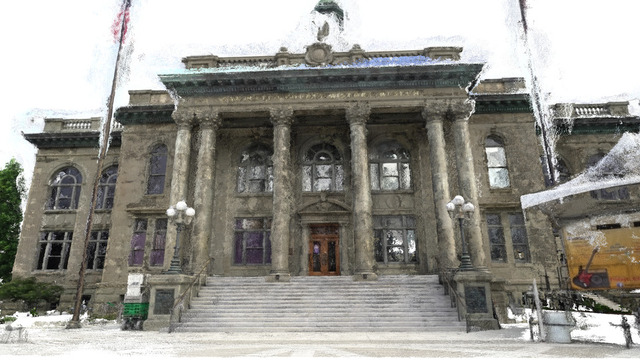}} &
		\raisebox{-.5\height}{\includegraphics[width=0.19\linewidth]{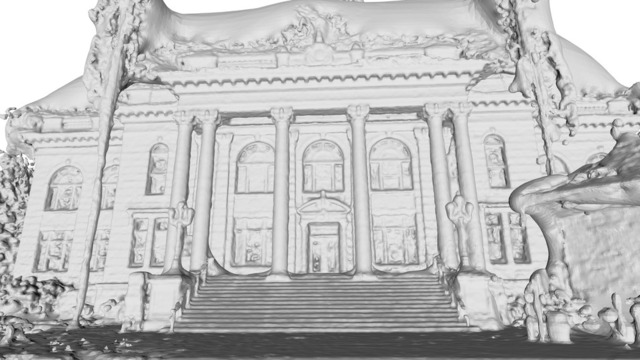}} &
		No Surface &%\includegraphics[width=0.19\linewidth]{image/tnttr/Courthouse/render_mesh_neus_nomask.jpg} &
		\raisebox{-.5\height}{\includegraphics[width=0.19\linewidth]{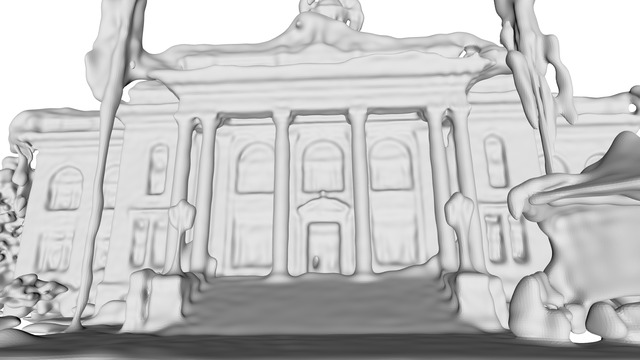}} &
		\raisebox{-.5\height}{\includegraphics[width=0.19\linewidth]{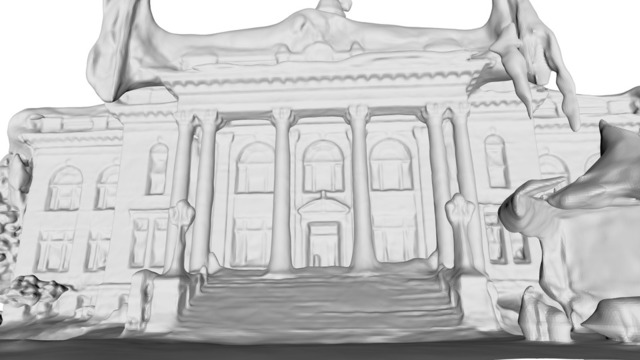}} \\
		
		\vspace{4pt}
		\rotatebox[origin=c]{90}{\small Ignatius} &
		\raisebox{-.5\height}{\includegraphics[width=0.19\linewidth,trim={150 0 150 50},clip]{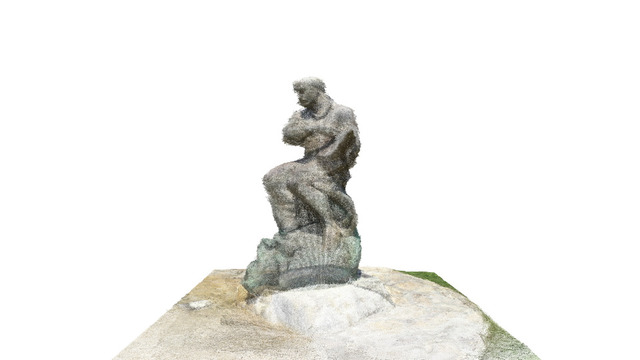}} &
		\raisebox{-.5\height}{\includegraphics[width=0.19\linewidth,trim={150 0 150 50},clip]{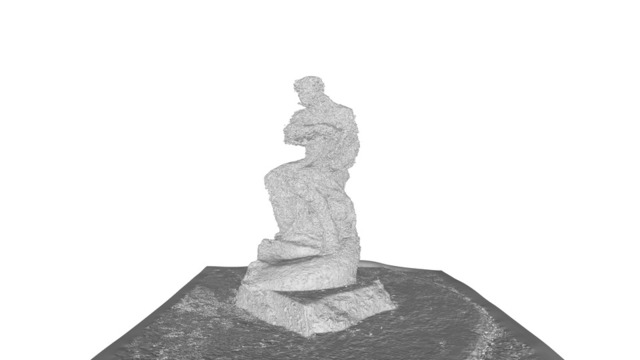}} &
		\raisebox{-.5\height}{\includegraphics[width=0.19\linewidth,trim={150 0 150 50},clip]{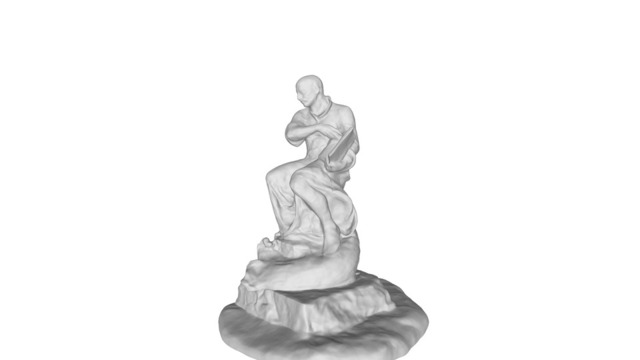}} &
		\raisebox{-.5\height}{\includegraphics[width=0.19\linewidth,trim={150 0 150 50},clip]{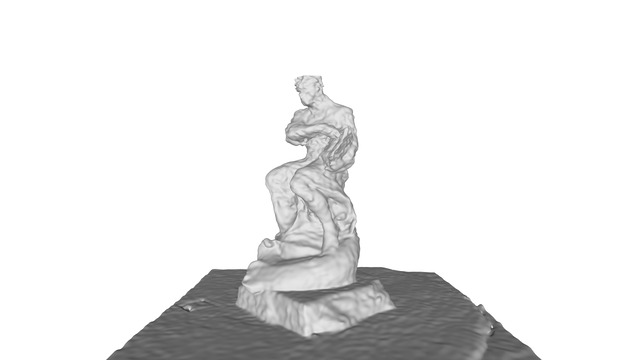}} &
		\raisebox{-.5\height}{\includegraphics[width=0.19\linewidth,trim={150 0 150 50},clip]{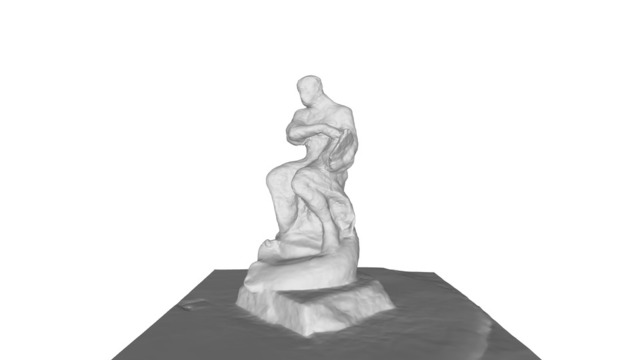}} \\
		& Point Cloud & sPSR & NeuS & SIREN & RegSDF (Ours)
	\end{tabular}
	\vspace{-3mm}
	\caption{Qualitative results on BlendedMVS \cite{yao2020blendedmvs} and Tanks and Temples \cite{knapitsch2017tanks} dataset.}
	\label{fig:bld_qual_supp}
	\vspace{-3mm}
\end{figure*}

\end{document}